\documentclass[single-column]{fairmeta}

\usepackage[most]{tcolorbox}

\usepackage{amsmath}
\usepackage{amssymb}
\usepackage[disable]{todonotes}
\usepackage{listings}
\usepackage{multirow}
\usepackage{makecell}
\usepackage{graphicx}
\usepackage{wrapfig}
\usepackage{float}
\usepackage{tabularx}
\usepackage{textcomp}
\usepackage{stfloats}
\usepackage{url}
\usepackage{verbatim}
\usepackage{titlesec}
\usepackage{tocloft}
\usepackage{adjustbox}
\usepackage{multirow}
\usepackage{pifont}
\usepackage{tikz}
\usepackage{comment}
\usepackage{amsmath,amssymb} % define this before the line numbering.
\usepackage{colortbl}  %
\usepackage{color}
\usepackage{booktabs} 
\usepackage{hyperref}
\usepackage{graphicx}    % 
\usepackage{subcaption} 
\usepackage{multirow} % 
\usepackage{booktabs} % 
\usepackage{subcaption} 
\RequirePackage{xspace}
\makeatletter
\DeclareRobustCommand\onedot{\futurelet\@let@token\@onedot}
\def\@onedot{\ifx\@let@token.\else.\null\fi\xspace}
\usepackage[most]{tcolorbox}
\usepackage{xcolor}
\usepackage{booktabs}
\usepackage{multirow}
\usepackage{makecell}
\usepackage{diagbox}
\usepackage{adjustbox}
\usepackage[table]{xcolor}
\usepackage{xltabular}
\usepackage{array}
\usepackage{threeparttable}

\definecolor{wmgreen}{RGB}{46,125,50}
\definecolor{wmorange}{RGB}{230,145,56}
\definecolor{wmgray}{RGB}{120,120,120}

\newcommand{\cmark}{\textcolor{wmgreen}{\ding{51}}}
\newcommand{\pmark}{--}
\newcommand{\xmark}{\textcolor{wmgray}{\ding{55}}}
\newcommand{\github}{\raisebox{-1.5pt}
{\includegraphics[height=1.05em]{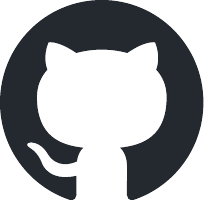}}\xspace}
\newcommand{\homepage}{\raisebox{-1.5pt}{\includegraphics[height=1.05em]{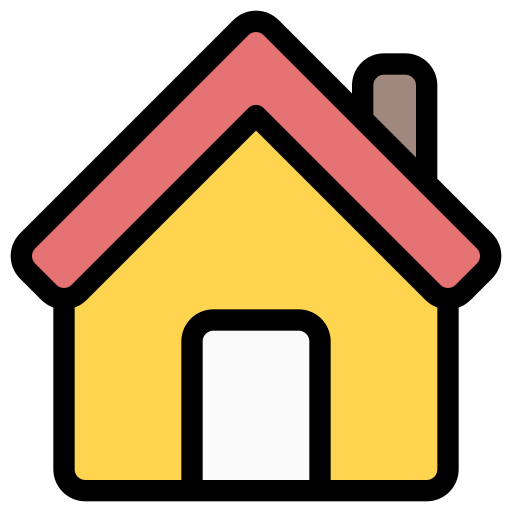}}\xspace}

\newcommand{\revise}[1]{\textcolor{black}{#1}}

\title{World Model for Robot Learning: A Comprehensive Survey}
\definecolor{myblue}{RGB}{233, 241, 249}
\definecolor{mygray}{RGB}{99, 110, 114}
\definecolor{myred}{RGB}{255, 118, 117}
\definecolor{myyellow}{RGB}{255, 234, 167}
\definecolor{mygreen}{RGB}{216, 226, 204}
\definecolor{mypurple}{RGB}{162, 155, 254}
\definecolor{mybrown}{RGB}{215, 190, 154}
\definecolor{myorange}{RGB}{255, 220, 190} 
\author[1,*,\dag]{Bohan~Hou}
\author[1,*]{Gen~Li}
\author[1,*]{Jindou~Jia}
\author[1,*]{Tuo~An}
\author[1,*]{Xinying~Guo}

\author[1]{Sicong~Leng}
\author[2]{Haoran~Geng}
\author[3]{Yanjie~Ze}
\author[4]{Tatsuya~Harada}
\author[5]{Philip~Torr}
\author[6]{Oier~Mees}
\author[7]{Marc~Pollefeys}
\author[8]{Zhuang Liu}
\author[3]{Jiajun~Wu}
\author[2]{Pieter~Abbeel}
\author[2]{Jitendra~Malik}
\author[9]{Yilun~Du}
\author[1,\dag]{Jianfei~Yang}

\small{
\affiliation[1]{Nanyang~Technological~University}
\affiliation[2]{University~of~California, Berkeley}
\affiliation[3]{Stanford~University}
\affiliation[4]{The~University~of~Tokyo}
\affiliation[5]{University~of~Oxford}
\affiliation[6]{Microsoft}
\affiliation[7]{ETH~Zurich}
\affiliation[8]{Princeton~University}
\affiliation[9]{Harvard~University}
}

\footnotesize{\contribution[*]{Equal Contribution (alphabetical order)}
\contribution[\dag]{Corresponding Author}}

\abstract{
World models, which are predictive representations of how environments evolve under actions, have become a central component of robot learning. They support policy learning, planning, simulation, evaluation, data generation, and have advanced rapidly with the rise of foundation models and large-scale video generation. However, the literature remains fragmented across architectures, functional roles, and embodied application domains. To address this gap, we present a comprehensive review of world models from a robot-learning perspective. We examine how world models are coupled with robot policies, how they serve as learned simulators for reinforcement learning and evaluation, and how robotic video world models have progressed from imagination-based generation to controllable, structured, and foundation-scale formulations. We further connect these ideas to navigation and autonomous driving, and summarize representative datasets, benchmarks, and evaluation protocols. Overall, this survey systematically reviews the rapidly growing literature on world models for robot learning, clarifies key paradigms and applications, and highlights major challenges and future directions for predictive modeling in embodied agents. To facilitate continued access to newly emerging works, benchmarks, and resources, we will maintain and regularly update the accompanying GitHub repository alongside this survey.
\begin{center}
    \renewcommand{\arraystretch}{1.2}
    \begin{tabular}{ll}
    
       \github  & \url{https://github.com/NTUMARS/Awesome-World-Model-for-Robotics-Policy}\\
       \homepage  & \url{https://ntumars.github.io/wm-robot-survey/} \\
    \end{tabular}
\end{center}

}

\correspondence{Jianfei Yang at \email{jianfei.yang@ntu.edu.sg};} 

\begin{document}

\maketitle

\section{Introduction}
Robotic policy learning is rapidly shifting from task-specific control pipelines toward foundation-model-driven embodied intelligence. Recent Vision-Language-Action (VLA)~\citep{rt2, openvla,pi_0,intelligence2025pi05,GR-1} policies aim to unify perception, language understanding, and control by mapping multimodal observations directly to robot actions, promising broad task generalization and flexible instruction following. Yet despite strong scaling trends~\citep{xiao2025worldenv,VLARFT,wmpo}, purely reactive VLA policies remain limited in complex physical environments, where they often struggle with long-horizon reasoning, temporal credit assignment, and robustness under compounding errors. A growing body of work argues that these limitations stem not only from insufficient action prediction capacity~\citep{dreamzero,rynnbrain}, but also from the lack of explicit predictive structure for anticipating how the world may evolve under the agent’s behavior. This has renewed interest in world models~\citep{40s,bryson1975applied,ha2018worldmodels}, predictive representations that capture environmental dynamics and enable reasoning about future states before acting.

The term \emph{world model}~\citep{40s, bryson1975applied, ha2018worldmodels} has a long intellectual lineage. At its core, it describes how a system or environment evolves from its current state under intervention or action, and in its most standard form can be viewed as a state-transition model that predicts the next state or a sequence of future states from the current state and action. Early ideas emerged in cognitive science during the 1960s~\citep{60s}, where internal models were proposed to support mental simulation, prediction, and planning. \revise{Similar ideas also appeared in control theory and model-based decision-making~\citep{conant1970every,bryson1975applied,richalet1978model}, and in classical robot planning, where internal models of geometry, constraints, and action consequences are used to support decision making before execution~\citep{lozano1983robot}.} In modern machine learning, the resurgence of world models is driven mainly by two lines of progress~\citep{ha2018worldmodels}: model-based reinforcement learning~\citep{nguyen1990improving,wovr,wmpo}, which uses learned dynamics for planning and policy improvement, and large-scale generative modeling~\citep{CosmosPredict,ctrlworld,rynnvla001,dreamgen}, especially video generation, which learns rich spatiotemporal regularities from large-scale visual or interaction data. Together, these developments make it increasingly plausible to learn predictive representations directly from pixels and reuse them for embodied decision making.

\begin{figure*}[t]
    \centering
    \includegraphics[width=\linewidth]{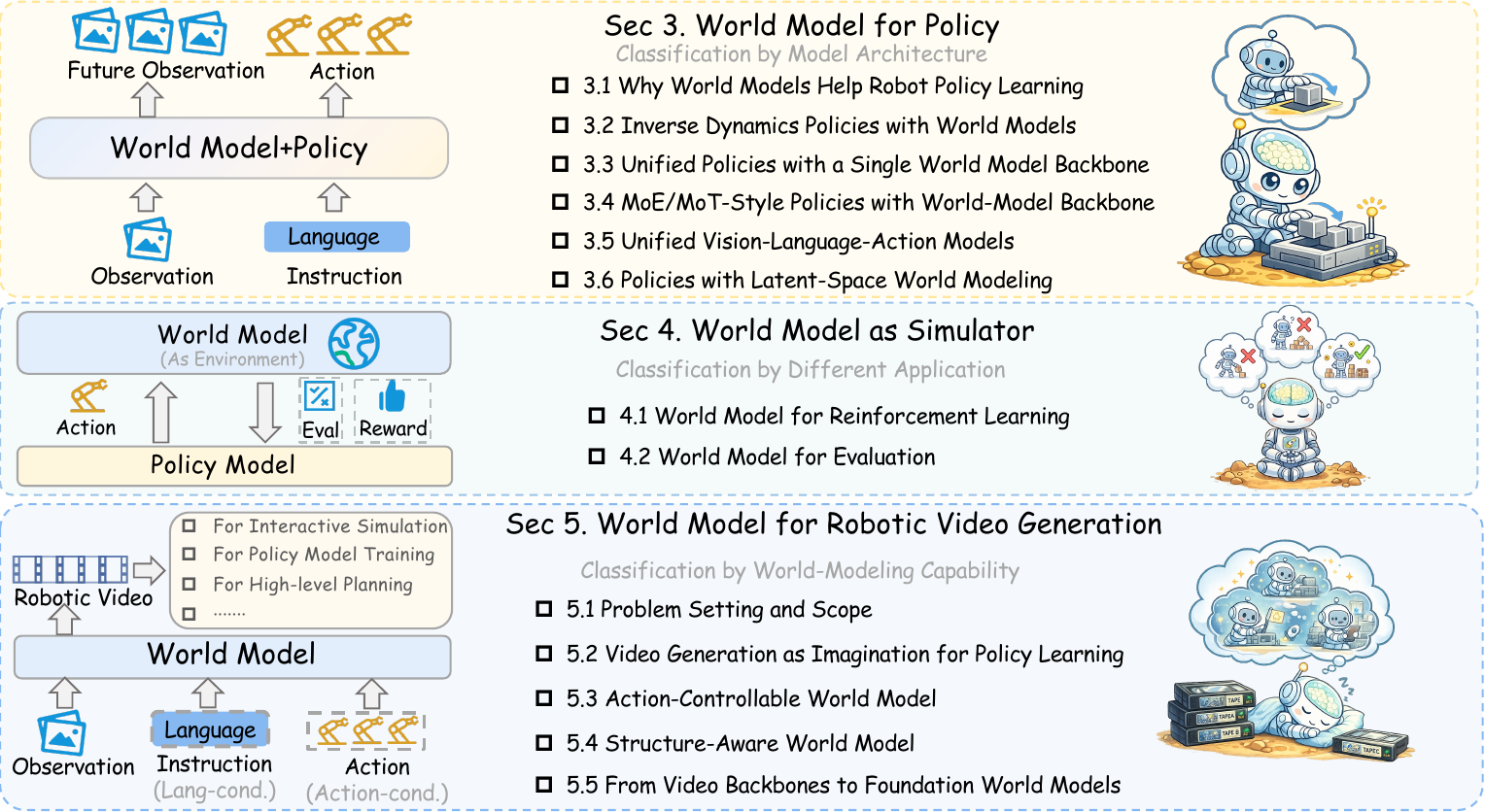}
    \caption{
    Overview of the organization of this survey.
    \textbf{Section 3} reviews how world models are coupled with robot policies from an architectural perspective.
    \textbf{Section 4} examines world models as simulators from an application perspective.
    \textbf{Section 5} focuses on robotic video world models and organizes the literature by world-modeling capability.
    }
    \label{fig:survey_roadmap}
\end{figure*}

In this survey, rather than enforcing a single narrow formal definition, we take a robot-learning-centered view of world models. Our focus is on how predictive models of future world evolution support robotic policy learning, planning, simulation, evaluation, and data generation.  Under this view, world models may support action selection through explicit rollout, future-conditioned action inference, or joint predictive-control modeling.  What unifies them is not a single factorization, but their role as predictive structures that make robot decision-making more informed and physically grounded. We also use the notion of action in a broad predictive-control sense: \textit{low-level motor commands} specify how the agent moves, while \textit{high-level language instructions} specify what the future should be realized. This perspective also distinguishes robotic world models from generic perceptual predictors: in embodied AI, predictive quality matters only insofar as it is useful for action. 
%Accordingly, an actionable world model should provide three core capabilities: \textbf{foresight}~\citep{tcidm,lingbotva,saydream,motus}, \textbf{imagination-driven planning}~\citep{CosmosPolicy}, and \textbf{data amplification}~\citep{dreamgen,CosmosPredict}. 
\revise{Accordingly, an actionable world model should provide three core capabilities: \textbf{foresight}~\citep{tcidm,lingbotva,saydream,motus}, i.e., anticipating future states or action consequences before execution; \textbf{imagination-driven planning}~\citep{CosmosPolicy}, i.e., using imagined rollouts to compare and select candidate behaviors; and \textbf{data amplification}~\citep{dreamgen,CosmosPredict}, i.e., synthesizing additional demonstrations or interaction trajectories to improve learning.}
These capabilities are especially important for embodied tasks such as manipulation, navigation, and driving, where success depends on reasoning about contact, dynamics, and other physical regularities that language-centric pretraining alone does not capture. In this sense, world models are not merely a generative enhancement, but a predictive bridge from semantic intent to physically realizable behavior.

Historically, the integration of world models into robotic policies has evolved along two directions: tighter coupling between predictive modeling and action generation~\citep{unipi,UVA,UWA}, and broader use of learned world models as simulators for validation, post-training, and reinforcement learning~\citep{xiao2025worldenv,VLARFT,diwa}. With the rise of foundation-scale video models~\citep{wan2025,CosmosPredict}, recent methods explore adapting large video generators into robot policies~\citep{UVA,UWA}, aiming to improve generalization and sample efficiency through future prediction~\citep{dreamgen}, while later systems move toward unified training and closed-loop co-optimization with VLA policies~\citep{rynnvla002}. In parallel, world models are increasingly used as controllable simulators for post-training and evaluation~\citep{wmpo,xiao2025worldenv}, highlighting that the key objective is not only to generate plausible futures, but to generate control-consistent futures that support decision-making.
\begin{figure*}[t]
    \centering
    \includegraphics[width=\linewidth]{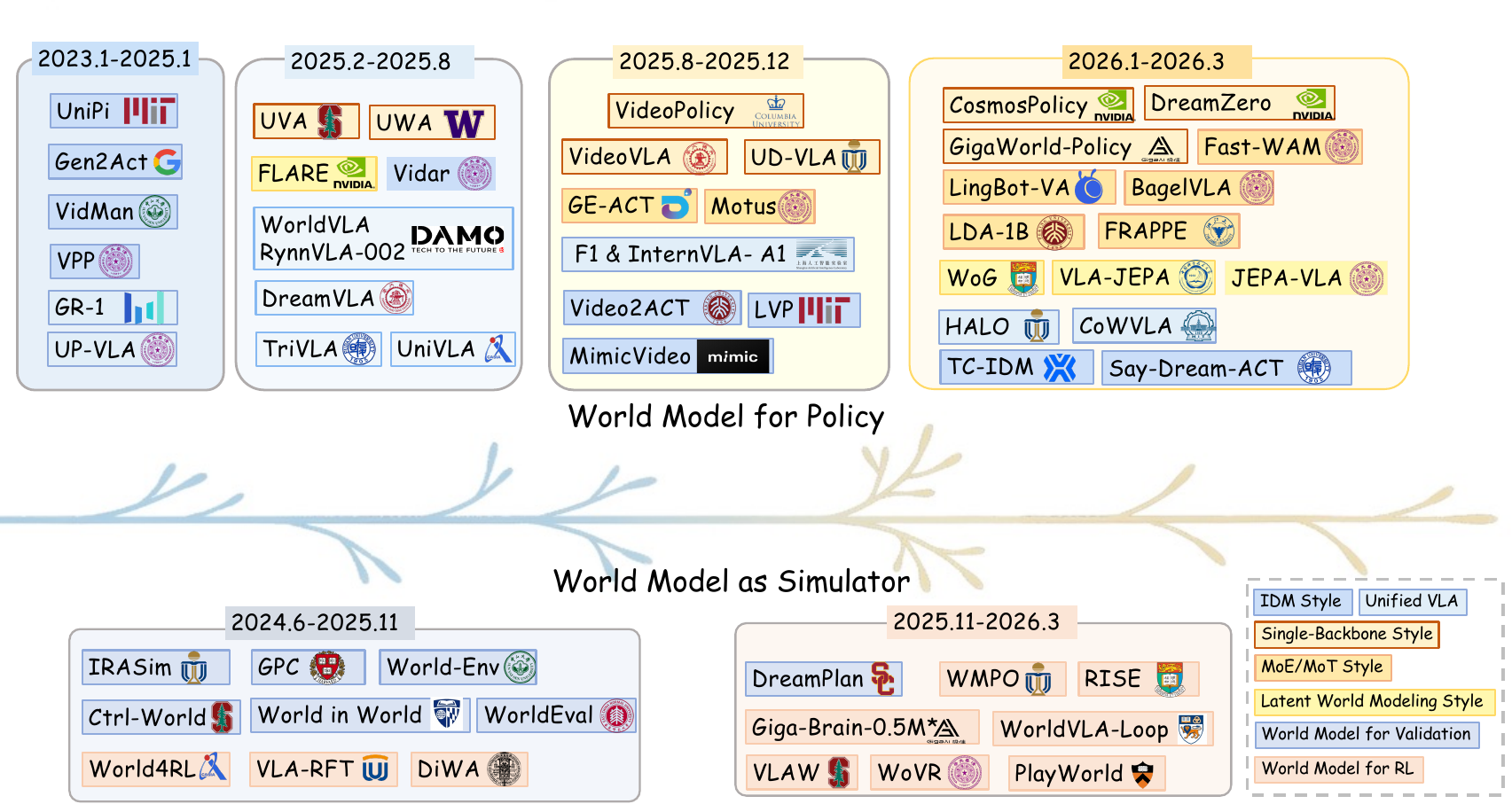}
    \caption{
    Temporal evolution of representative works on world models for robotic policy learning.
    The upper branch summarizes the progression of \textbf{world model for policy methods}, showing a trend from early decoupled video-generation-plus-IDM pipelines toward tighter integration through single-backbone, MoE/MoT, unified VLA, and latent world-modeling designs.
    The lower branch summarizes the progression of \textbf{world model as simulator methods}, where world models evolve from rollout-based validation and candidate evaluation into learned simulators for policy reinforcement learning, post-training, and co-evolving optimization.
    These trends should be understood as dominant directions rather than strictly sequential replacements.
    }
    \label{fig:wm-evolution}
\end{figure*}
Motivated by these trends, our survey differs from prior surveys~\citep{worldmodel-survey} in three main respects: it offers a more fine-grained view of major world-model paradigms, a more comprehensive analysis of their roles across policy learning, planning, simulation, evaluation, and video generation, and a clearer robotics-centered definition of world models in relation to VLA policies and robot learning. By emphasizing action-conditioned consistency, long-horizon reliability, and practical deployability, this survey aims to clarify when and why world models translate into measurable gains in real robotic behavior.

We first introduce background on world models, video generation, and VLA/policy models in Sec.~\ref{sec:background}. As summarized in Fig.~\ref{fig:survey_roadmap}, we then review world models for policy in Sec.~\ref{sec:policy}, world models as simulators in Sec.~\ref{sec:sim}, and robotic video world models in Sec.~\ref{sec:data_gen}. We further discuss broader embodied domains including navigation and autonomous driving in Sec.~\ref{sec:other_app}, and present benchmarks, datasets and results in Sec.~\ref{sec:benchmark}, before concluding with open challenges and future directions in Sec.~\ref{sec:challenges}. In particular, Sec.~\ref{sec:policy} first introduces a probabilistic lens that connects policy models, passive and controllable world models, and inverse-dynamics models as related queries of a shared predictive-control distribution.
 
\revise{Figure~\ref{fig:wm-evolution} highlights two closely related trends in the recent literature. On the policy side, early decoupled pipelines~\citep{vpp,unipi} remain an important line, while the design space has progressively expanded toward single-backbone~\citep{CosmosPolicy}, unified VLA~\citep{rynnvla002}, and latent world-modeling~\citep{wog} approaches with tighter integration between prediction and action generation.}
On the simulator side, their roles have expanded from validating or ranking candidate actions based on imagined futures to serving as learned environments for reinforcement learning, post-training, and even co-evolution with policies~\citep{VLARFT,guovlaw,worldvlaloop}. Taken together, these two trends indicate that world models are no longer used only as auxiliary predictors, but are increasingly integrated into the core learning and decision-making loop of robotic systems. To complement this survey, we will also continuously maintain and update the accompanying GitHub repository so that it remains aligned with the fast-moving progress of the field.

In summary, our main contributions are as follows:
\begin{itemize}
    \item We present a policy-centric survey of world models for robot learning, with a particular focus on how predictive models are coupled with VLA policies to support action generation, planning, simulation, evaluation, and data generation.
    \item We provide a more fine-grained taxonomy of the field by distinguishing major architectural paradigms and functional roles of world models, revealing important differences that are often overlooked in broader discussions.
    \item We offer a more comprehensive and clearly defined treatment of robotic world models by clarifying their relationship to robot learning, VLA policies, video generation, and simulator-style usage, and by summarizing representative benchmarks, datasets, and open challenges.
\end{itemize}
\section{Background}
\label{sec:background}
\subsection{World Model and Video Generation Model}
\todo[inline, color=mygreen]{Assigned to: Xinying Guo}
%\gxy{1st draft, do not read 11/03}

%\gxy{2nd draft, edited on 14/03}

%\hbh{modified 18/03 , ref from saining / lecun ,etc; align with Sec 3  }

%\hbh{we regard language as a high-level action , and video as a special state . the defeination of wm can be : a state transfer to another state by action. }

%\begin{figure}[t]
    %\centering
    %\includegraphics[width=0.95\linewidth]{figures/vla_wm_crop.pdf}
    %\caption{\revise{We'll update this figure soon }
    %Comparison between a VLA policy and a world model.
    %Left: a VLA policy predicts the next robot action from the current observation and language instruction.
    %Right: a world model predicts future observations from the current observation, sometimes conditioned on high-level task instruction and, when available, low-level action inputs.
    %In this sense, the policy focuses on \emph{what action to take next}, whereas the world model focuses on \emph{how the embodied scene evolves under instructions or actions}.
    %}
    %\label{fig:vla-vs-worldmodel}
%\end{figure}

To establish a precise vocabulary for the remainder of this survey, we first clarify two closely related concepts used throughout the paper. In recent embodied AI literature, the term world model has been used rather broadly, referring to latent dynamics models, future state predictors, video predictors, and even implicit predictive structures inside large policies. Since our focus is policy-centric rather than purely generative, we use these terms in a more precise and functional sense.

\subsubsection{World Model}

In this survey, we use the term world model in a robotics- and embodiment-centered sense, rather than in the broadest possible generative sense. Concretely, a world model refers to a predictive model of agent-environment dynamics that captures how a robotic or embodied system evolves under actions. In its most standard form, it models a state-transition process: given the current state or observation together with an action, it predicts the next state or a sequence of future states as illustrated in Fig.~\ref{fig:survey_roadmap} bottom.

Here we use the notion of action in a broad predictive-control sense. That is, both low-level motor commands and high-level language instructions are treated as \textit{actions}: the former are concrete physical actions executed by the agent, while the latter are high-level semantic actions that specify what the future should be realized. For notational consistency with the rest of this survey, we keep these two forms of action separate, denoting low-level physical actions by $a$ and high-level language or task actions by $l$. Under this convention, a general formulation can be written as
\begin{equation}
p(x_{t+1:t+H} \mid x_t, a_{t:t+H-1}, l),
\end{equation}
where $x_t$ denotes the modeled state at time $t$, $a_{t:t+H-1}$ denotes an action sequence over a horizon $H$, and $l$ denotes the high-level action specification, such as a language instruction or goal description. This formulation is intentionally agnostic to the choice of state space. What matters in our setting is whether the predicted futures are actionable for downstream embodied decision making.

Under this formulation, we use \emph{world model} in a functional sense to refer to predictive models whose outputs support policy-related computation, including control, planning, simulation, evaluation, and data generation. Its defining property is not merely to predict a plausible future, but to predict how the future changes under robot-relevant actions in a way that supports embodied decision making. This definition is therefore narrower than generic future prediction in computer vision: a model does not qualify as a world model in our sense simply because it generates plausible future images or videos. Rather, it must capture environment evolution in a form relevant to robot interaction and useful for downstream policy-related computation. In embodied control, the most important subclass is the action-conditioned world model, since visually plausible but action-inconsistent futures offer limited value for closed-loop decision making. Depending on the method, the modeled variable $x_t$ may be a visual observation, latent state, structured physical state, or even an abstract symbolic state used for planning~\citep{exopredicator,visualpredicator,pix2pred,exopredicator}, covering both classical latent dynamics models and newer generative predictive models for robot learning. In the symbolic case, the world model predicts transitions over predicates, object relations, affordances, or causal processes rather than over pixels~\citep{visualpredicator,pix2pred,exopredicator}.

In current embodied systems, however, the most common and scalable realization of state is precisely an observation stream, especially a visual observation sequence. For this reason, many practical world models in robotics are instantiated directly in visual observation space. Accordingly, although \emph{world model} is the more general concept, the concrete models of primary interest in this survey are predominantly visual world models, i.e., video generation models defined over future observations.

\subsubsection{Video Generation Model}

A video generation model predicts the future directly in image or video space. In the embodied setting, it can be written as
\begin{equation}
p(v_{t+1:t+H} \mid o_t, a_{t:t+H-1}, l),
\end{equation}
where $o_t$ denotes the current observation, it can represent observations from multiple perspectives, and $v_{t+1:t+H}$ denotes future frames or video segments. Compared with latent-state world models, this formulation preserves richer spatial, temporal, and interaction details, since the future is represented explicitly as visual evidence rather than abstract state variables. From the perspective above, such a model can be understood as a world model instantiated in visual observation space. Because visual observation is the most common form of state available to embodied agents, this visual instantiation is also the dominant one considered throughout this survey. This focus should not be read as assuming that pixel-level prediction is the optimal abstraction for control; rather, it reflects the prominence of video-based world models in the recent robot-learning literature.

This visual explicitness, however, also makes the modeling problem substantially more demanding. Beyond perceptual realism, an embodied video generation model must maintain temporal coherence, action consistency, physical plausibility, and long-horizon stability. Recent advances in large-scale video generative backbones have made such modeling increasingly viable in robotics \citep{yangcogvideox}. As a result, video generation models are no longer used only for passive visual continuation. They are increasingly adapted into action-conditioned predictive modules that support imagination-based supervision, controllable rollout, simulator construction, and synthetic data generation for robot learning \citep{liang2024dreamitate,zhou2024robodreamer,mimicvideo,Zhu_2025_ICCV,guo2026ctrlworld,huang2026vidworld,liao2026genie}.

\revise{Among them, action-conditioned video generation models occupy a particularly important place in embodied AI. Here, the notion of action should be understood broadly: conditioning may come from low-level continuous controls, but also from higher-level task or language descriptions that specify what the future should be realized. Under both forms, these models inherit the expressive power of video prediction while modeling how the visual future changes as a consequence of candidate actions. This makes them especially suitable for the policy-centric setting of this survey: they can serve not only as generators of plausible futures, but also as predictive substrates for control, planning, and policy improvement.} Therefore, unless otherwise specified, the world models discussed in the remainder of this survey are predominantly video-based world models, with special emphasis on the action-conditioned case.

\subsection{Robot Policy}
\todo[inline, color=myorange]{Assigned to: Jindou Jia}

%\jjd{drafted by jjd 02/25}

%\jjd{revised by jjd 02/28}

% State-of-the-art robot learning methods have shifted from analytical controllers to VLA models. Unlike traditional policies that map states directly to actions, the robot policy in VLA models leverages the reasoning capabilities of Vision-Language Models (VLMs) by co-fine-tuning them on robotic trajectory data.

\revise{State-of-the-art robot control methods have shifted from analytical controllers to end-to-end learning models~\citep{ai2025review}. Formally, the robot policy is a decision-making model that frames physical control as an action prediction task, mapping current environmental observations to future action trajectories. Here, we specifically focus on the imitation learning paradigm, where policies are trained to synthesize behaviors directly from expert demonstrations.}

\revise{Given the current observation $o_t$ (including visual and proprioceptive states) and an \textit{optional} language instruction $l$, the policy predicts future action sequences $a_{t+1:t+k}$. This process is typically modeled as the following conditional probability distribution:}
\begin{equation}
p(a_{t+1:t+k} | o_t, l).
\end{equation}
\revise{In practice, structuring predicted actions as temporal chunks with length $k$ has emerged as a predominant strategy to ensure temporal coherence and mitigate compounding errors~\citep{chi2023diffusion_RSS, zhao2023learning, wu2026vlanext}. From an architectural perspective, contemporary robot policies are primarily bifurcating into two paradigms: specialized visuomotor policies and generalist Vision-Language-Action (VLA) models. The former, represented by frameworks like Diffusion Policy~\citep{chi2023diffusion_RSS, chi2025diffusion, dasari2024ingredients}, focuses on training task-specific, often lightweight, end-to-end networks that leverage generative modeling to capture complex action distributions with high precision and low latency. Conversely, VLA models, pioneered by RT-2~\citep{rt2}, OpenVLA~\citep{openvla}, and $\pi_0$~\citep{pi_0}, are developed by fine-tuning large-scale Vision-Language Models (VLMs) on large scale robotic trajectory data~\citep{open_x_embodiment_rt_x_2023}, thereby inheriting the vast semantic knowledge and open-vocabulary reasoning capabilities of foundational models to achieve superior cross-task~\citep{octo_2024} and cross-embodiment generalization~\citep{Doshi24-crossformer}.}

\subsubsection{Visuomotor Policy}

\revise{Visuomotor policies establish a direct mapping from raw states to the action space, resulting in a generally lightweight yet generalization-bounded architecture. The most straightforward approach formulates this mapping as a \textit{regression} task~\citep{bain1995framework, osa2018algorithmic, zhao2023learning}. In this paradigm, neural networks encode the current observation and directly regress the continuous physical action values deterministically.}

\revise{To address the inherent multi-modality of human demonstrations, recent visuomotor policies have increasingly adopted \textit{generative} models. These approaches capture the full action distribution using generative techniques, such as Diffusion Policy~\citep{chi2023diffusion_RSS, chi2025diffusion} based on diffusion models~\citep{ddpm, song2020denoising}, and flow matching~\citep{zhang2024affordance, flowmatching, rectifiedflow}. By framing action prediction as a conditional generation process, these models can synthesize high-fidelity, multimodal action sequences starting from initial {Gaussian} noise. Furthermore, to enhance sampling efficiency and accelerate the generation process, recent advancements have explored replacing the standard Gaussian noise with more informative base distributions, such as visual representations~\citep{gao2025vita} and action history~\citep{jia2026action}. \citet{pan2026much} found that generative policies outperform regression by improving manifold adherence through stochasticity injection and supervised iterative computation.}

\subsubsection{Vision-Language-Action Policy}

\revise{To leverage the powerful reasoning capabilities of VLM, VLA models typically equip the pre-trained backbone with a dedicated action head, co-fine-tuning the entire framework on robotic trajectory data.
%, as shown in Fig.~\ref{fig:vla-vs-worldmodel}. 
In VLA models, action prediction mainly integrates discrete and continuous representation paradigms.} 

\revise{On one hand, \textit{discrete action tokenization} quantizes continuous actions into tokens that reside within the same vocabulary space as the language model, directly utilizing the next-token prediction ability of VLM~\citep{hybridvla}, successfully exemplified by RT-2~\citep{rt2} and OpenVLA~\citep{openvla}. While standard binning-based discretization can struggle with high-frequency control, FAST~\citep{pertsch25-fast} introduces Frequency-space Action Sequence Tokenization (FAST), which uses the discrete cosine transform (DCT) to compress action chunks into a dense token sequence. This enables autoregressive VLAs to handle highly dexterous tasks with the precision of generative models while significantly reducing training time. Its imitation-learning objective is identical to the standard negative log-likelihood loss.} 

\revise{On the other hand, to overcome quantization errors and maintain precision in high-frequency control, \textit{continuous action representation} has emerged as a promising alternative. This approach usually treats the action head as a conditional generator, learning a probabilistic generative model, such as diffusion models or flow matching, successfully exemplified by the $\pi$ family~\citep{pi_0, intelligence2025pi, intelligence2025pi05}. Instead of predicting deterministic values, these generative formulations model the full multimodal distribution of human demonstrations.}

\section{World Model for Policy}
\label{sec:policy}
The upper branch of Fig.~\ref{fig:wm-evolution} places existing policy-coupling paradigms in a broader temporal context. The field has gradually moved from decoupled predict-then-act pipelines to more unified and internalized forms of predictive control. Importantly, this progression should not be read as implying that video-pretrained backbones are inherently superior to VLM, latent, structured, or symbolic alternatives for control. Which predictive substrate is most effective remains an open empirical question; our focus here is to organize the rapidly growing family of methods that couple visual or video-based world-model priors with robot policies, while also highlighting latent and structured variants where they arise.

\todo[inline, color=myblue]{Assigned to: Bohan Hou, Gen Li, Jindou Jia}

\subsection{Why World Models Help Robot Policy Learning}
%\hbh{drafted by hbh 02/25}

Recent robotics policies increasingly incorporate world models, often instantiated as video generative models, because large-scale video pretraining may provide useful priors over temporal dynamics and physical regularities. Their benefit is not limited to predicting the future, but also lies in providing structured predictive representations that make action generation less ambiguous. By conditioning on anticipated outcomes rather than only the current observation, the policy gains longer-horizon foresight and a more informative basis for control.

This trend can be interpreted from a probabilistic perspective. Suppose the objective is to model the joint conditional distribution of future observations and future actions. Let \(o_t\) denote the current observation, \(a_t\) denote an action, and \(l\) denote the instruction (e.g., a language instruction or task specification). The distribution is expressed as
\begin{equation}
p(o_{t+1:t+k}, a_{t+1:t+k} \mid o_t, l).
\end{equation}
Then several seemingly distinct paradigms can be viewed as different marginals or conditionals of the same underlying predictive-control model:
\begin{align}
& \text{Policy Model:}\quad
p(a_{t+1:t+k} \mid o_t, l)
=
\int p(o_{t+1:t+k}, a_{t+1:t+k} \mid o_t, l)\, d_o, \\
& \text{Passive World Model:}\quad
p(o_{t+1:t+k} \mid o_t, l)
=
\int p(o_{t+1:t+k}, a_{t+1:t+k} \mid o_t, l)\, d_a, \\
& \text{Controllable World Model:}\quad
 p(o_{t+1:t+k} \mid o_t, a_{t+1:t+k}), \\
& \text{Inverse Dynamics Model:}\quad
 p(a_{t+1:t+k} \mid o_{t:t+k}).
\end{align}
In this sense, policy model, passive world model (video generation model), controllable world model and inverse dynamics model are not entirely separate abstractions; rather, they correspond to different ways of querying or factorizing the same idealized joint distribution. This also explains why world models and policies can be naturally coupled: a policy may use future observations generated by a world model as an intermediate latent variable, while an inverse-dynamics-style decoder can recover executable actions from such predicted futures.
\begin{table}[t]
\caption{Comparison of architectural paradigms for world-model-based policies in Sec.~\ref{sec:policy}.}
\centering
\footnotesize
\setlength{\tabcolsep}{3.2pt}

\renewcommand{\arraystretch}{1.08}
\begin{tabular}{cllcc}
\hline
\textbf{Paradigm} & \textbf{Representative Work} & \textbf{Future Generation at Inference} & \textbf{Backbone} & \textbf{Coupling Style} \\
\hline

\multirow{10}{*}{\centering IDM-style}
& UniPi~\citep{unipi} & Explicit video rollout & VGM & Decoupled \\
 
& VidMan~\citep{vidman} & Explicit video rollout & VGM & Decoupled \\
 
& Vidar~\citep{vidar} & Explicit video rollout & VGM & Decoupled \\
 
& Gen2Act~\citep{gen2act} & Explicit human-video rollout & VGM & Decoupled \\
 
& VPP~\citep{vpp} & Latent predictive features & VGM & Decoupled \\
 
& Video2Act~\citep{video2act} & Latent predictive features & VGM & Decoupled \\
 
& MimicVideo~\citep{mimicvideo} & Latent visual plan & VGM & Decoupled \\
 
& TC-IDM~\citep{tcidm} & Structured execution plan & VGM & Decoupled \\
 
& LVP~\citep{lvp} & Visual plan & VGM & Decoupled \\
 
& Say-Dream-ACT~\citep{saydream} & Video prompt & VGM & Decoupled \\
\hline

\multirow{8}{*}{\centering Single-backbone}
& UVA~\citep{UVA} & Joint latent prediction & VGM & Shared backbone \\
 
& UWA~\citep{UWA} & Joint diffusion process & VGM & Shared backbone \\
 
& VideoVLA~\citep{videovla} & Joint video rollout & VGM & Shared backbone \\
 
& VideoPolicy~\citep{VideoPolicy} & Video policy substrate & VGM & Shared backbone \\
 
& Cosmos Policy~\citep{CosmosPolicy} & Parallel action/state/value outputs & VGM & Shared backbone \\
 
& DreamZero~\citep{dreamzero} & Chunk-wise joint rollout & VGM & Shared backbone \\
 
& UD-VLA~\citep{udvla} & Synchronous denoising & VGM & Shared backbone \\
 
& GigaWorld-Policy~\citep{gigaworldpolicy} & Optional visual branch & VGM & Shared backbone \\
\hline

\multirow{7}{*}{\centering MoE/MoT}
& GE-Act~\citep{gen2act} & Latent visual guidance & VGM & Expert fusion \\
 
& Motus~\citep{motus} & Expert rollout & VGM & MoT fusion \\
 
& LingBot-VA~\citep{lingbotva} & Visual predictive context & VGM & MoT fusion \\
 
& BagelVLA~\citep{bagelvla} & Single-step visual foresight & VGM & MoT fusion \\
 
& Fast-WAM~\citep{fastwam} & Train-time video, test-time skipped & VGM & MoT fusion \\
 
& LDA-1B~\citep{lda} & Latent dynamics only & VGM & Expert fusion \\
 
& FRAPPE~\citep{frappe} & Latent representation alignment & VGM & Parallel experts \\

& DiT4DiT~\citep{dit4dit} & Latent video guidance & VGM & Expert fusion \\
\hline

\multirow{10}{*}{\centering Unified VLA}
& GR-1~\citep{GR-1} & Future image prediction & UMM & Joint co-training \\
 
& UP-VLA~\citep{upvla} & Future image prediction & UMM & Joint co-training \\
 
& WorldVLA~\citep{rynnvla002} & Future image (mainly train-time) & UMM & Joint co-training \\
 
& DreamVLA~\citep{dreamvla} & Structured world knowledge & UMM & Joint co-training \\
 
& UniVLA~\citep{unifiedvla}& Latent world modeling & UMM & Joint co-training \\
 
& CoWVLA~\citep{cowvla} & Latent dynamics & UMM & Joint co-training \\
 
& F1~\citep{f1vla} & Visual foresight & UMM & Unified MoT \\
 
& InternVLA-A1~\citep{cai2026internvla} & Latent foresight & UMM & Unified MoT \\
 
& HALO~\citep{halo} & Visual subgoal prediction & UMM & Unified multi-expert \\
 
& TriVLA~\citep{TriVLA} & Episodic dynamics & UMM & Multi-system \\
\hline

\multirow{4}{*}{\centering Latent-space WM}
& FLARE~\citep{flare} & Latent alignment & MLLM & Latent internalization \\
 
& VLA-JEPA~\citep{vlajepa} & Latent target prediction & MLLM & Latent internalization \\
 
& JEPA-VLA~\citep{jepavla} & Predictive embeddings & MLLM & Latent internalization \\
 
& WoG~\citep{wog} & Future condition only & MLLM & Latent internalization \\

& DIAL~\citep{dial} & Latent visual foresight & MLLM & Latent internalization \\ 
\hline

\end{tabular}
\label{tab:sec3_architectural_summary}
\end{table}

Therefore, integrating a world model into policy learning can be viewed more generally as introducing predictive structure into action generation. Instead of learning a direct, monolithic mapping from the current observation to actions, the model reasons about future observations as auxiliary predictive variables that inform or constrain action selection. In some formulations, the model first predicts a plausible future and then decodes actions conditioned on that future. In others, candidate actions are generated first and then evaluated or regularized through predicted future outcomes. More unified approaches model observations and actions jointly within a shared generative process.

In practice, incorporating such predictive structure can provide a useful inductive bias for control, especially when robot action data are limited and large-scale predictive pretraining is available~\citep{dreamgen}. Motivated by this perspective, we organize this section from an architectural viewpoint, categorizing world-model-based policy methods according to how predictive generation interacts with action production, ranging from decoupled pipelines to tightly integrated end-to-end formulations. Table~\ref{tab:sec3_architectural_summary} summarizes these paradigms from a comparative architectural perspective, highlighting their representative methods, whether future generation remains active at inference time, the underlying backbone family, and the style of coupling between world modeling and action generation. This architectural progression also echoes broader trends in foundation models, where the design space has expanded from modular pipelines to include shared backbones, expert-coupled architectures, and latent forms of capability internalization.%Table~\ref{tab:sec3_architectural_summary} summarizes these paradigms by comparing their representative works, inference-time future generation, backbone family, and coupling style.

\subsection{Inverse Dynamics Policies with World Models}

\label{subsec:wm_idm_policy}
%\hbh{drafted by hbh 02/26}

\begin{figure}[t]
    \centering
    \includegraphics[width=\linewidth]{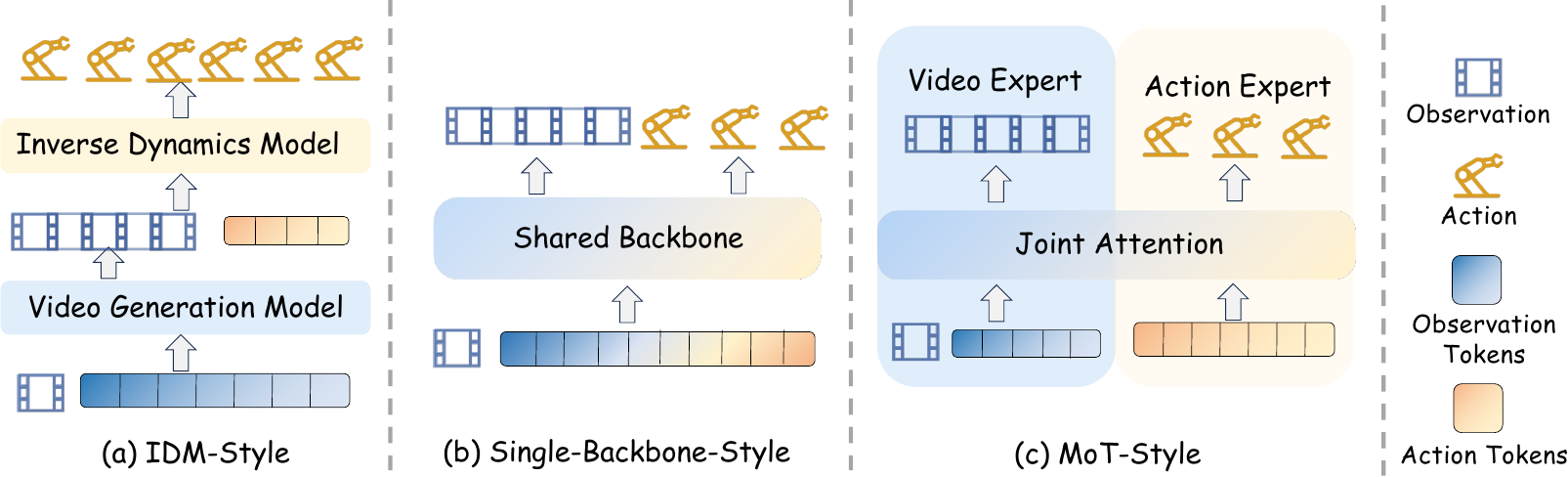}
    \caption{
    Representative architectural paradigms for using world models as policies.
    \textbf{(a) IDM-style.}
    A video generation model first predicts future observations, and an inverse dynamics model then recovers actions from the predicted visual trajectory, yielding a decoupled predict-then-act pipeline.
    \textbf{(b) Single-backbone-style.}
    Observation and action tokens are processed within a unified shared backbone, so future prediction and action generation are modeled jointly in a common latent space.
    \textbf{(c) MoT-style.}
    Video and action experts remain partially specialized, while cross-modal interaction is achieved through shared joint attention, enabling deeper coupling between world modeling and policy generation.
    }
    \label{fig:wm-policy-paradigms}
\end{figure}

A representative line of work incorporates world models into robot control through a decoupled design, in which future prediction and action generation are realized by two distinct modules. The central idea is to first use a world model, most commonly an image or video generative model, to predict a task-conditioned future observation sequence (or its latent representation), and then train a separate policy module to infer executable actions from the current observation together with the predicted future. Unlike unified end-to-end policies that jointly model perception, prediction, and control within a single backbone, this paradigm preserves an explicit functional separation: the world model provides a structured hypothesis of ``what should happen next,'' while the policy translates such anticipated futures into low-level actions. As illustrated in Fig.~\ref{fig:wm-policy-paradigms}(a), this family adopts a decoupled predict-then-act pipeline: a world model first produces future observations or their predictive representations, and a separate inverse-dynamics-style policy then maps these anticipated futures to executable actions.

This family of methods first constructs a predicted future trajectory:
\begin{equation}
\hat{\mathbf{o}}_{t+1:t+H} = \mathcal{W}(o_t, l),
\end{equation}
or, more generally, a future latent representation:
\begin{equation}
\hat{\mathbf{z}}_{t+1:t+H} = \mathcal{W}\!\left(\mathrm{E}_{\mathrm{img}}(o_t), \mathrm{E}_{\mathrm{text}}(l)\right),
\end{equation}
where $\mathcal{W}$ denotes the world model and $H$ is the prediction horizon. The policy is then conditioned on both the current observation and the generated future
\begin{equation}
\pi(a_{t+1,t+H'} \mid o_t, l)
= P\!\left(
a_t \,\middle|\,
\mathrm{E}_{\mathrm{img}}(o_t),
\mathrm{E}_{\mathrm{text}}(l),
\Phi(\hat{\mathbf{o}}_{t+1:t+H})
\right),
\end{equation}
or equivalently in latent form
\begin{equation}
\pi(a_{t+1,t+H'} \mid o_t, l)
= P\!\left(
a_t \,\middle|\,
\mathrm{E}_{\mathrm{img}}(o_t),
\mathrm{E}_{\mathrm{text}}(l),
\hat{\mathbf{z}}_{t+1:t+H}
\right),
\end{equation}

where $\Phi(\cdot)$ denotes a feature extractor over predicted future observations, and $H'$ denotes the action chunk size. From a control perspective, this formulation is inverse-dynamics-style: rather than inferring actions solely from the present state, the policy leverages an anticipated state transition or future evolution signal, thereby reducing ambiguity in action generation.

Historically, early works in this line established the basic decoupled paradigm of predict first, then act. UniPi~\citep{unipi} is a representative early example: it uses a task-conditioned world model to generate future video trajectories, and then trains a separate inverse dynamics model to derive an action representation by comparing adjacent frames. 
Subsequent methods mainly advance this paradigm by progressively redesigning what form of future representation is exposed to the policy. Early visual extensions, such as VidMan~\citep{vidman} and Vidar~\citep{vidar}, retain the canonical two-stage recipe while introducing masked inverse dynamics to emphasize action-relevant regions. Gen2Act~\citep{gen2act} follows a similar decoupled design, but conditions execution on generated human videos rather than robot-centric future rollouts. Besides, works such as VPP~\citep{vpp} and Video2Act~\citep{video2act} represent a closely related variant that moves away from explicit pixel-space rollout and instead treats the video world model as a source of compact predictive representations. Rather than decoding actions from fully rendered future frames, these methods extract control-relevant features from the latent space of a pretrained video diffusion model and inject them into a separate action head, yielding a more compact and stable interface between prediction and control. \revise{A related direction is explored by V2A~\citep{luo2025grounding}, which further grounds generated video states into action through goal-conditioned exploration: instead of learning a direct inverse-dynamics decoder from predicted futures, it treats synthesized video states as visual goals and learns a goal-conditioned policy through hindsight-style self-exploration.}
%In particular, Video2Act explicitly refines such representations into spatial and motion-aware cues and feeds them to a diffusion-based action decoder in an asynchronous dual-system manner, further illustrating how video-world-model priors can be converted into efficient, policy-facing conditions rather than treated as full generative rollouts.

Moving toward tighter representations, MimicVideo~\citep{mimicvideo} replaces explicit video prediction with partially denoised latent visual plans, yielding a more compact and control-aligned interface. TC-IDM~\citep{tcidm} and LVP~\citep{lvp} push this abstraction further by translating generated futures into execution-oriented intermediates, such as tool-centric geometric trajectories or retargetable visual plans. From a prompting perspective, Say-Dream-ACT~\citep{saydream} uses generated video plans not as explicit inverse-dynamics targets, but as in-context visual guidance for a separate action policy. Taken together, these methods reveal a common trend within this family: the predicted future gradually shifts from raw pixel-space rollout to increasingly structured, compact, and execution-friendly representations, while the world model and policy remain architecturally decoupled.

\revise{A related and complementary direction introduces more structured geometric intermediates into this decoupled pipeline. Rather than using generated or demonstrated videos only as raw visual futures, these methods further extract 3D-aware motion structure from video and use it as a more action-relevant predictive prior. In this sense, the key interface is still visually grounded, but the future is represented in a more structured form, such as dense correspondences, 3D trajectories, motion fields, or actionable 3D flow. Representative examples include AVDC~\citep{ko2024learning}, which recovers actions from synthesized videos through dense correspondences; VidBot~\citep{vidbot}, which extracts 3D hand trajectories and interaction cues from human videos; Object-centric 3D Motion Field~\citep{3dmf}, which represents actions through object-centric 3D motion structure similar to Hind4sight-Net~\citep{nematollahi2020hindsight}; and NovaFlow~\citep{li2025novaflow}, which distills generated videos into actionable 3D object flow for downstream execution. These works can be viewed as a structured extension of the decoupled paradigm: the world model remains visually grounded, but 3D representations serve as an intermediate structured prior that makes downstream action recovery more direct and robust.}

A defining feature of this family is its architectural decoupling: the predictive model is typically pretrained first and then frozen, lightly adapted, or connected to a separate policy head, rather than optimized jointly with action generation. This separation brings modularity, reusable video priors, and interpretable future prediction, but also limits performance by the fidelity and controllability of the generated future, and can accumulate error when visually plausible predictions are not action-consistent. Even so, this paradigm remains one of the earliest and most influential routes by which world models became directly useful for robot policy learning, and it naturally motivates the tighter couplings introduced in later video--action architectures.
%\todo[inline , color=myblue]{ dual-system-planner vla}

\subsection{Unified Policies with a Single World Model Backbone }
\label{subsec:unified_wm_policy}
%\hbh{drafted by hbh 03/04}
Different from the decoupled inverse-dynamics-style pipeline above, a more tightly coupled line of work uses a single generative backbone to jointly model future visual evolution and future actions. Figure~\ref{fig:wm-policy-paradigms}(b) summarizes this shift visually: instead of passing predictions from a world model to a downstream policy module, observation and action tokens are processed within one shared backbone, so future modeling and action generation are coupled inside the same generative process. A more fundamental motivation behind this design is not merely that video models can ``imagine'' future observations, but that pretrained video-generative backbones are optimized for temporally predictive modeling. In contrast to many VLM backbones~\citep{openvla,pi_0}, which are primarily pretrained through image--text or vision--language alignment objectives and therefore emphasize semantic correspondence, video generation models are trained to model temporally ordered observations and may encode useful priors over motion continuity, temporal causality, and approximate physical dynamics. When action generation is embedded into the same denoising or generative process that models future world evolution, the policy may therefore benefit from a backbone already biased toward propagating constraints across time. However, whether video-pretrained backbones are consistently superior to matched-scale VLM backbones for robotic control remains an open empirical question; current results should be viewed as suggestive evidence for a promising inductive bias rather than a definitive architectural conclusion.

At a high level, this family replaces the two-stage factorization of ``predict first, then act'' with a unified multimodal generative objective. Let $\mathbf{x}=[z^{v};z^{a}]$ denote the concatenation of future visual and action representations. A shared backbone $f_{\theta}$ is trained on corrupted inputs $\tilde{\mathbf{x}}_{\tau}$ under conditioning $(o_t,l)$
\begin{equation}
\hat{y} = f_{\theta}(\tilde{\mathbf{x}}_{\tau}, o_t, l, \tau), \qquad \mathbf{x}=[z^{v};z^{a}],
\end{equation}
where $\tau$ means denoising steps, with a generic unified objective
\begin{equation}
\mathcal{L}_{\mathrm{unified}} = \mathbb{E}\big[\ell(\hat{y}, y)\big],
\end{equation}
where the exact target depends on the specific instantiation: $y$ may correspond to diffusion noise in continuous denoising models, a velocity field in flow-matching variants, or masked tokens in discrete denoising formulations.

Representative early unified designs such as UVA~\citep{UVA}, UWA~\citep{UWA}, and later VideoVLA~\citep{videovla} already make this perspective explicit. Rather than preserving a modular ``world model + policy head'' decomposition, they treat control as a direct interface to a unified predictive generator. UVA learns a joint video--action latent space and supervises both modalities jointly, while retaining efficient deployment through lightweight modality-specific decoding heads that allow policy inference to bypass explicit video generation. UWA pushes this coupling further into the diffusion process itself by integrating video and action diffusion within a single transformer under modality-specific timesteps, and can be queried as a policy by marginalizing out the visual future through timestep control. Building on this unified view, VideoVLA shifts the emphasis toward directly converting a pretrained video generator into a robotic control model: it extends a Video Diffusion Transformer into a Video-Action Diffusion Transformer that jointly predicts future visual outcomes and action sequences, thereby making the pretrained video model itself the backbone of the policy. A closely related perspective is taken by VideoPolicy~\citep{VideoPolicy}, which treats video generation as the primary policy substrate and reduces action prediction to a lightweight interface layered on top of the generated behavior.

Subsequent methods tighten this coupling further by minimizing the representational gap between visual prediction and control. Cosmos Policy~\citep{CosmosPolicy} is a particularly direct realization of this idea: it keeps the pretrained video diffusion architecture essentially unchanged and encodes robot actions, future states, and values as additional latent ``frames'' within the original diffusion sequence. At inference time, these outputs need not all be used symmetrically: in direct policy mode, only the action output is required for execution, whereas in planning mode the future-state and value predictions can be used to rank candidate trajectories. DreamZero~\citep{dreamzero} follows the same end-to-end philosophy with an autoregressive flow-matching video-action DiT, but performs closed-loop chunk-wise joint denoising rather than free-running long-horizon rollout, thereby limiting compounding error while preserving tight video--action alignment. UD-VLA~\citep{udvla} extends the same principle to a discrete multimodal setting, coupling future-image tokens and action tokens within a single synchronous denoising trajectory while introducing dedicated test-time efficiency techniques. GigaWorld-Policy~\citep{gigaworldpolicy} provides a more explicitly action-centered variant: it jointly optimizes future action prediction and action-conditioned future video generation within a single shared transformer stack, while using a causal design that makes the visual branch optional at inference time.

Ultimately, the key difference across these unified methods is not whether they all render full future videos online, but how much of the visual branch remains active during control. Some preserve explicit future prediction for consistency or planning, whereas others retain the benefits of joint training while marginalizing, truncating, or partially discarding the visual branch for efficiency. In all cases, unlike the decoupled methods above, the world model is not treated as a separate upstream module consumed by a downstream policy. Instead, world modeling and policy learning are collapsed into a single generative process, providing one possible route for injecting spatiotemporal priors from large-scale video pretraining into control.

\subsection{MoE/MoT-Style Policies with Expert World-Model Backbones}
\label{subsec:moe_mot_video_backbone}
%\hbh{drafted by hbh 03/04}

Compared with the single-backbone generators above, a related but architecturally distinct line of work preserves explicit specialization by maintaining separate expert streams for video prediction, action generation, and sometimes language or scene understanding. Rather than collapsing all modalities into one shared diffusion backbone, these methods adopt MoE/MoT-style~\citep{mot} or multi-branch designs, where modality-specific experts interact through shared attention, cross-attention, or interleaved autoregressive sequences. As illustrated in Fig.~\ref{fig:wm-policy-paradigms}(c), unlike single-backbone models, they retain separate video and action experts while coupling them through repeated interaction. The motivation remains to transfer the spatiotemporal and physical priors of pretrained video diffusion models~\citep{wan2025,CosmosPredict} into control, but under a different architectural assumption: full parameter sharing is not always optimal, since video prediction and action generation have different temporal frequencies, representational scales, and optimization requirements. In this sense, these models resemble expertized VLA designs such as $\pi_0$~\citep{pi_0} and $\pi_{0.5}$~\citep{intelligence2025pi05}, except that their backbone is not primarily a static semantic encoder, but a temporally predictive video generator whose representations may contain useful cues about motion continuity, temporal causality, and approximate physical dynamics.

At a high level, these approaches can be viewed as learning a coupled predictive-control mapping with specialized experts:
\begin{equation}
\left(\mathbf{h}^{v}_{\ell+1},\, \mathbf{h}^{a}_{\ell+1}\right)
=
\mathcal{F}^{\mathrm{mix}}_{\ell}\!\left(\mathbf{h}^{v}_{\ell},\, \mathbf{h}^{a}_{\ell};\, o_t,\, l\right),
\end{equation}
where \(\ell\) indexes the layer, and $\mathcal{F}^{\mathrm{mix}}_{\ell}$ denotes a layerwise interaction operator, such as joint attention, cross-attention, or shared-attention fusion~\citep{unidiffuser}, that couples a video expert and an action expert while preserving their distinct parameterization. Under this view, the video branch serves as a temporally predictive latent stream, and the policy is obtained by repeatedly injecting this foresight into the action branch, rather than by decoding actions from an entirely separate downstream head.

Within this family, one common pattern is parallel expert coupling, where a pretrained video diffusion backbone is paired with a lighter action branch. GE-Act~\citep{liao2026genie} follows this pattern by introducing a parallel flow-matching action pathway alongside a pretrained video diffusion world model, using deep cross-attention to inject visual latent features into action generation. Here, the video branch provides predictive world-state structure, while the action branch translates it into executable control without requiring full video rendering online. Earlier instantiations of this paradigm leveraged image-editing diffusion models to predict subgoals for a goal-conditioned policy to follow~\citep{black2023zero,hatch2024videoglue}.

A second and more explicit pattern is Mixture-of-Transformers-based deep interaction, in which multiple experts are retained throughout the network and fused repeatedly via shared attention. Motus~\citep{motus}, LingBot-VA~\citep{lingbotva}, BagelVLA~\citep{bagelvla} and more recently DiT4DiT~\citep{dit4dit} are representative examples. Motus formulates the design most directly as a Mixture-of-Transformers with dedicated experts for understanding, video generation, and action. LingBot-VA pushes this idea toward causal world modeling by interleaving video and action tokens into a shared autoregressive sequence and using a dual-stream MoT with shared attention, turning imagined future states into a context for action refinement. BagelVLA extends the same intuition to longer-horizon manipulation, interleaving linguistic planning, visual forecasting, and action generation within one execution loop; its Residual Flow Guidance further makes visual foresight practical through single-step denoising~\citep{instaflow} rather than full video rollout. DiT4DiT~\citep{dit4dit} follows the same intuition by coupled architecture, using intermediate denoising features from the video branch to guide action prediction. Fast-WAM~\citep{fastwam} can be viewed as a hybrid point in this family: it adopts a shared-attention Mixture-of-Transformers backbone with coupled video and action branches, yet concludes that the main benefit may come more from video co-training during training than from explicit future imagination at inference time. Across these variants, the video branch is increasingly treated not as an output to be faithfully rendered, but as a predictive latent process whose hidden states guide action generation.

A third pattern is latent-space expertization, which shifts world modeling from pixel space to structured latent dynamics while retaining specialized multimodal branches. LDA-1B~\citep{lda} represents this direction by moving visual forecasting into a DINO~\citep{dino} latent space and coupling visual and action experts through shared self-attention inside a multimodal diffusion transformer. FRAPPE~\citep{frappe} follows a related philosophy from the perspective of future-representation alignment: instead of reconstructing future observations, it trains multiple parallel expert streams with separate adapters and aligns them to visual foundation models in latent space. Although more training-oriented than the explicitly architected MoT models above, it reflects the same underlying idea that deeply coupled specialized predictive streams can improve world-aware action generation.

Taken together, these methods bridge the gap between detached modular pipelines and fully unified single-backbone generators. They embed world modeling directly into the policy while preserving architectural specialization. Specifically, video diffusion models provide predictive foresight, while MoE/MoT mechanisms translate this into action without losing modality-specific structures. Compared to single-backbone approaches, the key distinction is architectural: both aim to couple future prediction with action generation, but these methods achieve it through deeply interacting specialized experts rather than full parameter sharing.

\begin{figure}[t]
    \centering
    \includegraphics[width=0.9\linewidth]{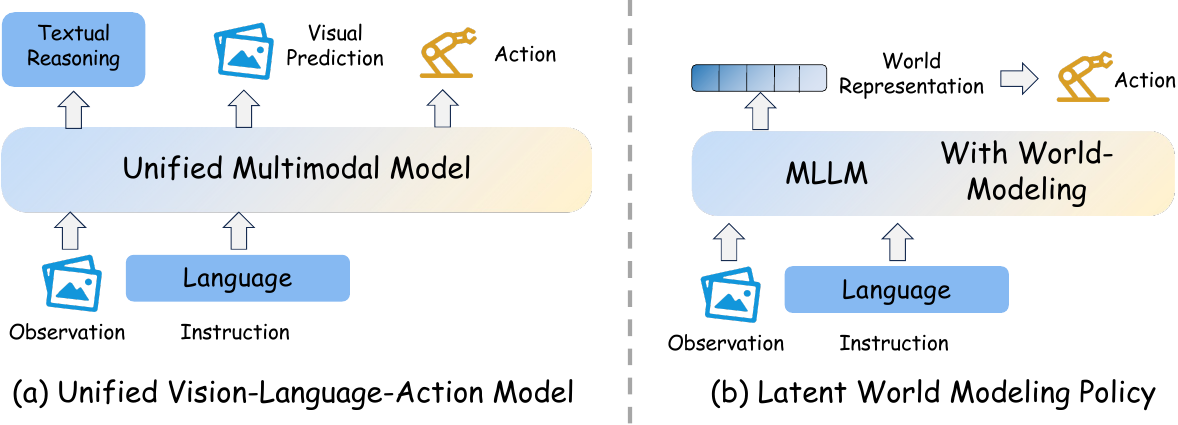}
    \caption{
    Two MLLM-based routes for internalizing world modeling into policy learning.
    \textbf{(a) Unified Vision-Language-Action model.}
    A unified multimodal model jointly processes observation and language, and is trained to produce action together with auxiliary future-oriented outputs such as textual reasoning or visual prediction.
    \textbf{(b) Latent-space world modeling for VLA.}
    Instead of explicitly predicting future images, the model internalizes future dynamics as a compact world representation or world modeling inside the same MLLM backbone, and then maps this latent predictive knowledge to action.
    }
    \label{fig:umm-vs-latentwm-vla}
\end{figure}
\subsection{Unified Vision-Language-Action Models}
\label{subsec:unified_vla}
Unified VLA models provide another route to world model as policy. Although they do not always employ an explicit video world model, they still learn future-oriented predictive structure within the same multimodal policy backbone, for example through future-image prediction, visual foresight, or structured world knowledge. As shown in Fig.~\ref{fig:umm-vs-latentwm-vla}(a), this family differs from the preceding video-backbone paradigms in that future modeling is internalized inside a unified VLA architecture rather than introduced through a separate predictive module.

One important subclass performs explicit future-state prediction. These methods directly predict future images, either a single frame or a short sequence, as part of the unified training objective. GR-1~\citep{GR-1} is an early representative that jointly predicts actions and future images within a single GPT-style transformer. UP-VLA~\citep{upvla} follows a similar strategy, using future-image prediction to improve both action generation and visual generalization. WorldVLA~\citep{rynnvla002} further unifies action and image understanding and generation in one autoregressive framework, while using future-image prediction mainly as a joint training signal rather than a mandatory inference-time output.

A second subclass replaces pixel-level prediction with implicit or latent future modeling. Instead of forecasting future frames directly, these methods predict compact future-aware representations that are more tightly aligned with action. DreamVLA~\citep{dreamvla} predicts structured world knowledge, including dynamic, spatial, and semantic cues, to support inverse dynamics modeling. UniVLA~\citep{unifiedvla} incorporates world modeling during post-training over a native multimodal tokenization framework, allowing the model to absorb causal dynamics from large-scale video data without introducing a separate external world model. CoWVLA~\citep{cowvla} pushes this direction further by modeling latent motion and compact future visual targets instead of reconstructing redundant future frames.

A third subclass consists of \emph{multi-expert or multi-system unified models}, which remain unified at the training and task level but preserve explicit functional specialization inside the architecture. This category includes F1~\citep{f1vla}, InternVLA-A1~\citep{cai2026internvla}, HALO~\citep{halo}, and TriVLA~\citep{TriVLA}. Although these methods also adopt expertized or MoT-style designs, their predictive branch is better understood as visual foresight or subgoal generation within a unified VLA framework, rather than as a native video-backbone world model. F1 predicts future visual states as planning targets within a Mixture-of-Transformers architecture. InternVLA-A1 extends this design with lightweight latent visual foresight and joint optimization of foresight prediction and action generation. HALO pushes the predictive branch toward visual subgoal prediction and embodied reasoning, while TriVLA organizes grounding, episodic dynamics perception, and control as coordinated subsystems.

Taken together, unified VLA models extend the notion of world model as policy beyond explicit video generation. Some do so through direct future-image prediction, others through compact latent or semantic world knowledge, and still others through unified multi-expert systems with explicit foresight modules. Across these variations, the shared principle is that action generation is no longer treated as a purely reactive mapping from the current observation, but is jointly trained with an internal predictive objective that captures future state evolution or its compact surrogate. Relative to the preceding subsections, the key distinction is therefore not whether the model contains an explicit standalone world model, but whether future-oriented predictive modeling is internalized within the same multimodal policy backbone.

\subsection{Policies with Latent-Space World Modeling }
\label{subsec:latent_world_model_vla}
%\hbh{drafted by hbh 03/04}

A further route to world model as policy is defined by methods that internalize future prediction entirely in representation space, without relying on explicit image or video generation. Rather than synthesizing future observations, these approaches construct predictive latent targets, future-aware embeddings, or compact control conditions, and couple them with action generation within the same policy framework. In this context, world modeling is realized not as visual reconstruction, but as learning a future-aware representation that captures how the environment may evolve in a form directly useful for control. Such methods therefore retain the core benefit of world modeling by injecting predictive structure into action generation, while avoiding the computational overhead and redundancy of explicit generative decoding. Conceptually, this direction is related to the JEPA family~\citep{jepa,vjepa2}, which models prediction in embedding space rather than pixels, but the focus here is not JEPA itself; rather, it is the emergence of VLA methods that turn this representation-space predictive principle into a practical mechanism for policy learning. Figure~\ref{fig:umm-vs-latentwm-vla}(b) illustrates this more internalized variant. Here, the backbone is again typically MLLM-based rather than video-DiT-based, but future modeling is absorbed more deeply into latent world representations or parameterized world knowledge, so that action generation is guided by internal predictive structure without requiring explicit future-image decoding.

Chronologically, FLARE~\citep{flare} is an early representative of this direction. It introduces ``Future Latent Representation Alignment'', aligning hidden features of the action denoising network with latent embeddings of future observations, so that the policy can implicitly anticipate future states without explicitly generating them. VLA-JEPA~\citep{vlajepa} makes this more explicit by adopting a JEPA-style pretraining objective for VLA: its key design is leakage-free state prediction, where future frames are used only to produce latent targets for supervision, encouraging the model to learn action-relevant state transitions in latent space rather than shortcutting through pixel variation. JEPA-VLA~\citep{jepavla} takes a complementary route: instead of adding an explicit latent prediction head, it argues that predictive embeddings learned by video JEPA models, especially V-JEPA 2~\citep{vjepa2}, already provide stronger policy priors than static visual representations, and therefore adapts these predictive embeddings as a better backbone for existing VLA models. Most recently, WoG~\citep{wog} moves world modeling into the condition space of action generation: rather than predicting future images or generic future latents, it learns to predict compact future-oriented conditions together with actions, so that the model directly forecasts the part of future information that is most useful for precise control. DIAL~\citep{dial} provides a closely related recent example by decoupling high-level intent from low-level action through latent world modeling, using latent visual foresight in the VLM feature space as a structured bottleneck for downstream action generation.

Beyond neural latent representations, a related but more classical non-pixel abstraction appears in symbolic or planner-facing world models. Unlike the neural policy backbones reviewed above, these methods usually externalize world modeling as an abstract transition model over predicates, object relations, affordances, operators, or causal processes, which is then queried by a symbolic or task-and-motion planner to produce high-level skill sequences~\citep{symbolic_operators_tamp,lamp_symbolicwm,visualpredicator,pix2pred,exopredicator}. We include this line as a complementary perspective to emphasize that useful world models may not depend on predicting pixels; they can also capture abstract logic, object relations, causal regularities, and symbolic dynamics for planning and control.

Taken together, this subsection highlights a non-pixel route for world-model-based policy learning. The main line is latent-space world modeling, where the policy avoids explicit future-image or video decoding while still internalizing future dynamics for action generation. The  symbolic planner  examples above further reinforce the same broader point: control-relevant prediction can be expressed through compact latent or abstract variables when they provide a more direct interface to action.
\section{World Model as Simulator}
\label{sec:sim}

Beyond serving as a predictive module for conditioning, planning, or internal supervision, world models can also be used more directly as interactive simulators. In this paradigm, the value of a world model lies not only in its ability to model future evolution, but in its ability to stand in for the environment itself~\citep{xiao2025worldenv,VLARFT,wmpo,team2025evaluating}: given the current observation, task instruction, and candidate actions, the model can roll out future states, provide feedback signals, and support downstream decision making through imagined interaction. This makes world model as simulator a particularly direct and practical extension of world modeling for embodied intelligence. 

This direction is especially appealing for \revise{visualmotor policies } because conventional reinforcement learning on physical robots is often slow, expensive, difficult to reset, and potentially unsafe, while pure imitation learning remains limited by demonstration quality and cannot easily learn from failures. Recent work therefore replaces costly real-world interaction with learned simulators built from world models, enabling policy improvement through imagined rollouts rather than repeated physical trial-and-error~\citep{wu2023daydreamer}. In frameworks such as World-Env~\citep{xiao2025worldenv}, VLA-RFT~\citep{VLARFT}, and WMPO~\citep{wmpo}, the world model is explicitly used as a low-cost, controllable virtual environment for post-training, substantially reducing the dependence on real interaction while improving data efficiency and robustness.

At the same time, the simulator view provides benefits beyond reinforcement training. Because a world model can roll out action-conditioned future states, it can also expose verifiable signals from predicted trajectories, such as reward-like feedback~\citep{xiao2025worldenv,VLARFT}, task completion cues, or rollout consistency, which are useful not only for policy optimization but also for evaluation, ranking, and test-time decision making. This is already reflected in systems that augment the learned simulator with reward or termination feedback, and it naturally extends to rollout-based assessment of candidate behaviors. Here, the same predictive ability that supports imagined training also becomes a mechanism for judging whether a candidate action sequence is likely to succeed before it is executed. 

\begin{figure}[t]
    \centering
    \includegraphics[width=0.9\linewidth]{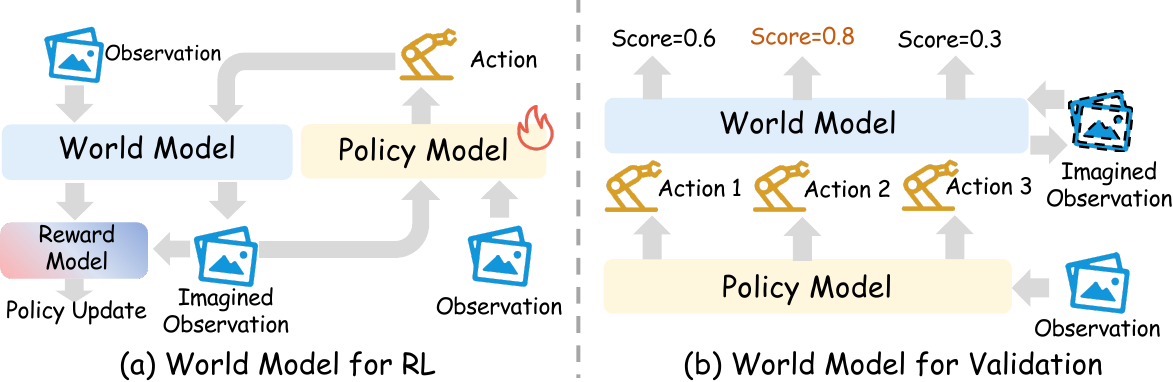}
    \caption{
    Two uses of world models for policy learning.
    \textbf{(a)} In the reinforcement-learning setting, the world model serves as a learned simulator that produces imagined transitions for policy improvement.
    \textbf{(b)} In the validation setting, the world model scores imagined consequences of candidate actions to support decision-time selection.
    }
    \label{fig:wm-rl-vs-validation}
\end{figure}

For this reason, we organize this section around two complementary uses of the simulator paradigm. As summarized in Fig.~\ref{fig:wm-rl-vs-validation}, world models can support policy learning in at least two distinct functional roles: as learned simulators for reinforcement learning, and as evaluators for decision-time validation.  We first discuss world model for reinforcement training, where the learned simulator is used to replace physical interaction and support policy improvement through imagined rollouts. We then discuss world model for evaluation, where the same simulator capability is used to verify, rank, or assess candidate behaviors through predictive rollout and future-state feedback. %Recent work~\citep{} also shows that once world models are treated as simulators, their reliability becomes a central concern: long-horizon hallucination and model error can directly corrupt optimization and evaluation signals, motivating a growing line of work on controllable rollouts, simulator--policy alignment, and closed-loop co-improvement. 
%\subsection{World Model For Reinforcement Training}
\todo[inline, color=myblue]{Assigned to: Bohan Hou, Xinying Guo,  jindou Jia}

\subsection{World Model for Reinforcement Learning}
\label{subsec:wm_for_rl}
%\hbh{drafted by hbh 03/05}

A further role of world models in embodied learning is to serve as interactive simulators for reinforcement post-training. Different from the preceding paradigms, where the world model mainly provides predictive conditioning, planning cues, or internal supervision, the methods in this subsection directly use the world model as a learned environment in which a Vision-Language-Action (VLA) policy can roll out trajectories, receive rewards, and improve through imagined interaction. In this setting, the world model is no longer merely a predictor of plausible futures; it becomes the medium through which reinforcement learning is carried out. 
At a high level, these methods optimize a policy inside a learned simulator
\begin{equation}
(\hat{o}_{t+1}, \hat{r}_t, \hat{d}_t) \sim p_{\phi}(\cdot \mid o_{\le t}, a_{\le t}, l),
\end{equation}
where the world model $p_{\phi}$ provides imagined transitions, and optionally rewards and termination signals. The policy is then improved from imagined rollouts by maximizing expected return,
\begin{equation}
J(\theta) = \mathbb{E}_{\hat{\tau}\sim(\pi_{\theta}, p_{\phi})}\!\left[\sum_{t} \gamma^t \hat{r}_t\right],
\end{equation}
or, in practice, through a GRPO-style policy optimization objective,
\begin{equation}
\mathcal{L}_{\mathrm{RL}}(\theta)
=
-\mathbb{E}_{t}\Big[
\min\big(
r_t(\theta)\hat{A}_t,\,
\mathrm{clip}(r_t(\theta),1-\epsilon,1+\epsilon)\hat{A}_t
\big)
\Big],
\qquad
r_t(\theta)=\frac{\pi_{\theta}(a_t\mid s_t)}{\pi_{\theta_{\mathrm{old}}}(a_t\mid s_t)},
\end{equation}
together with task-specific variants adapted to chunked or flow-based action heads. Under this shared view, the common goal is to replace expensive physical interaction with reinforcement learning inside a learned world simulator. 

Within this first-level paradigm, early works such as \revise{UniSim~\citep{unisim}}, World-Env~\citep{xiao2025worldenv} and VLA-RFT~\citep{VLARFT} establish the basic recipe of reinforcement learning in a learned simulator by coupling action-conditioned world simulation with reward generation. DiWA~\citep{diwa} shows that a frozen world model learned from large-scale play data can already support fully offline adaptation of diffusion policies, while World4RL~\citep{world4rl} extends this idea to higher-fidelity manipulation refinement through a diffusion world model for end-to-end imagined policy optimization.

Subsequent works make this paradigm increasingly compatible with modern VLA architectures and larger embodied datasets. World-Gymnast~\citep{quevedo2025worldgym} demonstrates that RL inside a video world model can outperform both supervised finetuning and software simulators. PlayWorld~\citep{playworld} learns robot world models from autonomous play and shows that reinforcement learning in the learned simulator improves downstream real-world performance. RehearseVLA~\citep{rehearse} adapts the same principle to VLA post-training with a physically consistent world simulator and an instant reflector for reward and termination feedback. In parallel, WMPO~\citep{wmpo} emphasizes pixel-space imagination and on-policy GRPO, ProphRL~\citep{prophrl} adapts RL updates to flow-based action heads through FA-GRPO and FlowScale, RISE~\citep{RISE} augments the simulator with compositional dynamics and progress-value estimation, and GigaBrain-0.5M$^*$~\citep{gigabrain} scales world-model-based RL to pretrained VLA adaptation. Despite these differences, all of these methods treat the world model primarily as the environment in which policy optimization takes place.

A second-level development explicitly recognizes that the learned simulator is itself imperfect and must be improved together with the policy. World-VLA-Loop~\citep{worldvlaloop}, VLAW~\citep{guovlaw}, and WoVR~\citep{wovr} exemplify this shift in different ways. World-VLA-Loop~\citep{worldvlaloop} jointly predicts future observations and rewards, and closes the loop by using policy failure rollouts to refine the simulator. VLAW~\citep{guovlaw} follows an iterative repair-and-improve strategy, alternating between real-world data for simulator refinement and synthetic data for VLA improvement. WoVR~\citep{wovr} pushes this direction further by treating simulator reliability as the central bottleneck, introducing controllable action-conditioned video modeling, Keyframe-Initialized Rollouts, and explicit world-model--policy co-evolution:
\begin{equation}
\phi^{k+1} \leftarrow \mathrm{UpdateWM}\!\left(\phi^k, D_{\mathrm{real}} \cup D_{\mathrm{policy}}(\pi_{\theta^k})\right), \qquad
\theta^{k+1} \leftarrow \mathrm{UpdatePolicy}\!\left(\theta^k, \hat{D}(\phi^{k+1})\right),
\end{equation}
where policy rollouts refine the world model, and the improved world model in turn produces better imagined data for subsequent policy updates. In this sense, the focus moves beyond reinforcement learning in a world model toward reinforcement learning with a world model whose fidelity, action-following precision, and rollout reliability must themselves be continuously improved.

Taken together, this subsection reveals a clear progression in the role of world models for policy improvement. The first-level paradigm treats world models as learned simulators for reinforcement training, differing mainly in reward design, rollout representation, and optimization compatibility. The second-level paradigm further recognizes that imagined reinforcement learning is only as effective as the simulator is reliable, and therefore introduces simulator refinement, rollout regulation, and policy--world-model co-evolution as integral parts of the loop.

\subsection{World Model for Evaluation}
\todo[inline, color=mygreen]{Assigned to: Xinying Guo, Bohan Hou, jindou Jia}

%\gxy{1st draft, do not read}

%\gxy{2nd draft, edited on 15/03}

%\hbh{revised }
Beyond serving as a learned simulator for reinforcement post-training, a world model can also evaluate candidate behaviors before execution. Here, the goal is not to improve a policy through repeated imagined interaction, but to estimate which candidate action sequence, policy, or checkpoint is most likely to succeed in the real world. As illustrated in Fig.~\ref{fig:wm-rl-vs-validation}(b), the world model supports decision-time selection by scoring or verifying imagined consequences of candidate actions. Given the current observation, task instruction, and one or more candidate actions, it rolls out predicted futures and uses them for ranking, rejection, or safety filtering. In this sense, the evaluator role is a natural extension of the simulator view: once a world model can stand in for the environment, it can be used not only for training in imagination, but also for judging what the policy should do next.

One direct form of evaluation is rollout-based candidate assessment. Here, the policy proposes multiple action sequences, the world model predicts their future outcomes, and the system selects the candidate with the most favorable imagined consequence. GPC~\citep{gpc} is a particularly clean example: rather than retraining the policy, it augments a frozen generative robot policy at deployment with an action-conditioned world model and uses predictive look-ahead to rank and refine candidate actions online. IRASim~\citep{Zhu_2025_ICCV} similarly demonstrates model-based planning by simulating multiple candidate trajectories and selecting the one with the highest predicted value. World-in-World~\citep{zhang2025world} extends this idea to closed-loop planning, where candidate plans are rolled out in imagination, evaluated by a revision policy, and revised before execution. DreamPlan~\citep{jia2026dreamplanefficientreinforcementfinetuning} turns the same evaluator logic into a training signal by constructing preference pairs over candidate actions from world-model rollouts. Across these methods, the world model acts as a decision-time or near-decision-time selector that converts imagined futures into action choice.

 Beyond simple ranking of discrete candidates, a more active paradigm treats the world model as the transition dynamics for Model Predictive Control (MPC). In this setting, the system does not merely select from a few pre-defined actions but actively optimizes an action sequence within the world model's imagined trajectories to minimize a cost function. Works such as TD-MPC2~\citep{hansen2024tdmpc2} and LeWorldModel~\citep{leworldmodel} demonstrate that latent-space MPC can significantly enhance the long-horizon reasoning of embodied agents. By performing gradient-based planning through the world model, the agent can discover complex strategies that are not explicitly present in the training demonstrations. This synergy effectively transforms the world model from a passive judge of actions into an active navigational map for continuous control optimization.

A second, more explicit form is to use the world model itself as a policy evaluator. \revise{A recent large-scale example is Evaluating Gemini Robotics Policies in a Veo World Simulator~\cite{veorobotics2025}, which uses a video world simulator for offline policy evaluation, OOD testing, and safety probing.} WorldEval~\citep{li2025worldeval} is the clearest example: it studies whether a world model can serve as a scalable proxy for real-world policy evaluation, ranking different robot policies and even different checkpoints of the same policy entirely in imagination, while also functioning as a safety detector. The same role appears at the benchmark level in WorldArena~\citep{shang2026worldarena}, which explicitly identifies policy evaluation as a core downstream use of embodied world models.

A third form arises when the simulator is equipped with explicit feedback heads that convert imagined rollouts into assessment signals. World-Env~\citep{xiao2025worldenv} augments the simulator with continuous reward prediction and action termination prediction. VLA-RFT~\citep{VLARFT} uses verified rewards computed on imagined trajectories inside a controllable world simulator. World-VLA-Loop~\citep{worldvlaloop} jointly predicts future observations and reward signals in a state-aware video world model. RISE~\citep{RISE} makes this evaluator role even more explicit by introducing a progress value model that scores imagined outcomes according to task advancement. In these systems, imagined rollouts are not only a source of synthetic training data, but also a basis for deciding whether a behavior is promising, complete, or worth executing.

A related but lighter-weight perspective appears in latent-space predictive world models, especially the JEPA line. Rather than generating explicit pixel-space futures for ranking candidate actions, these methods perform prediction and planning in embedding space. V-JEPA 2~\citep{vjepa2} and V-JEPA 2.1~\citep{vjepa2_1} are representative examples, with the latter further showing that a latent action-conditioned world model can support zero-shot robot planning with image goals. More recently, LeWorldModel~\citep{leworldmodel} pushes this direction toward a simpler and faster end-to-end JEPA formulation, while also showing that latent predictive models can detect physically implausible events. At present, however, these methods are better viewed as an adjacent direction for predictive planning and plausibility checking than as fully developed policy evaluators in the broader embodied-control sense considered here.

This evaluator perspective also clarifies why action faithfulness and rollout reliability matter so much. An evaluator is useful only if its imagined future preserves the causal consequences of candidate actions. Ctrl-World~\citep{guo2026ctrlworld} makes this connection explicit by showing that action-faithful rollouts can support policy evaluation in imagination. At the same time, WoVR~\citep{wovr} highlights an important caveat: hallucination and long-horizon error do not merely reduce visual quality, but can directly corrupt the assessment signal itself. For evaluation, realism alone is therefore insufficient; what matters is whether the rollout remains reliable enough to support ranking, selection, and rejection in a way that tracks real-world execution.

\revise{Taken together, these works reveal a clear broadening of the simulator paradigm. In embodied robot learning, a world model is no longer only a low-cost environment for reinforcement training; it is increasingly also used as an evaluator that can compare policies, score candidate behaviors, detect likely failures, and support both decision-time action selection and offline policy assessment~\citep{li2025worldeval,veorobotics2025}. This shift is conceptually important for the remainder of the survey, because it shows that the value of a world model lies not only in generating future trajectories, but in generating trajectories that are trustworthy enough to support policy evaluation and action choice.}

\section{World Model for Robotic Video Generation}
\label{sec:data_gen}
\todo[inline, color=mygreen]{Assigned to: Xinying Guo}

% \jjd{A recent related work: https://sites.google.com/view/vlaw-arxiv, which tries to iteratively co-improve Vision-Language-Action Policy and World Model, where World Model is used for Data Generation. FYI.}

% \hbh{
% just view our repo to get the newest
% }

%\gxy{1st draft, do not read first 11/03}

%\gxy{2nd draft, edited on 14/03}

%\hbh{revised 03.19}

%\gxy{revised 01/04}

\begin{figure}[!t]
    \centering
    \includegraphics[width=\linewidth]{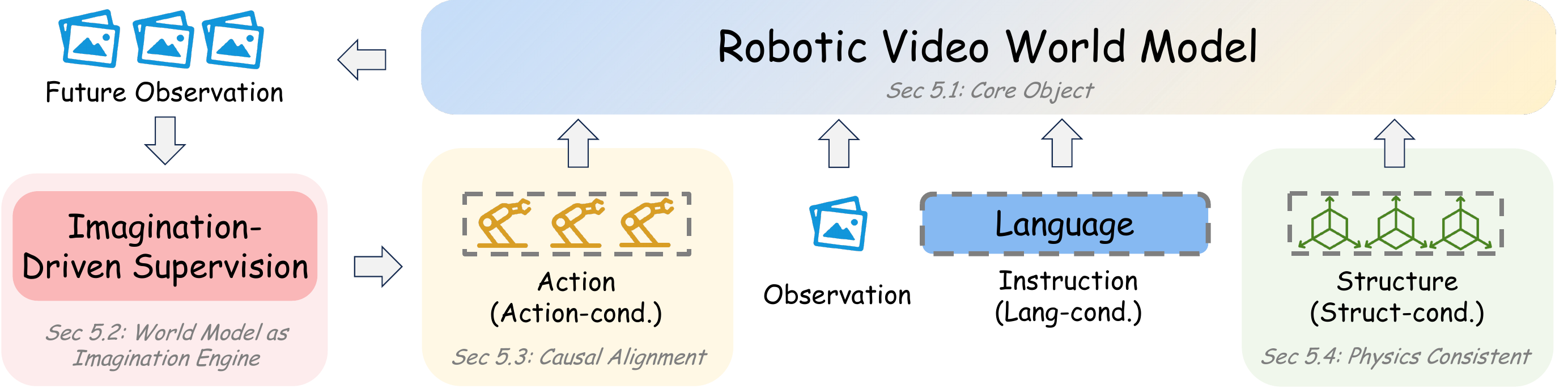}
    \caption{
    Unified view of robotic video world models.
    \textbf{Section 5.1} defines the core object as a robotic video world model that predicts future observations in visual space.
    Building on this core, \textbf{Section 5.2} uses the predicted future as an imagination engine for supervision,
    \textbf{Section 5.3} introduces action conditioning to improve causal alignment and controllability,
    and \textbf{Section 5.4} further incorporates structure priors to enhance physical and interaction consistency.
    The resulting future observation is therefore not only visually plausible, but increasingly actionable for downstream robot learning and decision making.
    Finally, \textbf{Section 5.5} highlights the broader transition from task-specific video prediction to a scalable and reusable world-model interface built on strong video priors.
    }
    \label{fig:section5}
\end{figure}

% in preamble:
% \usepackage{booktabs}
% \usepackage{amssymb}

% Requires: \usepackage{booktabs,multirow}

% Requires:
% \usepackage{booktabs}
% \usepackage{multirow}
% \usepackage{makecell}
% \usepackage{pifont}
% \newcommand{\cmark}{\ding{51}}
% \newcommand{\xmark}{\ding{55}} % optional, not used here

\begin{table*}[t]
\centering
\footnotesize
\setlength{\tabcolsep}{3.2pt}
\renewcommand{\arraystretch}{1.06}
\caption{
Comparison of representative methods in Sec.~\ref{sec:data_gen}, grouped by the four capability regimes discussed in Sec.~5. 
Checkboxes indicate whether a feature is explicitly supported or emphasized by the original paper. 
Here, \textit{Foundation-scale} is reserved for methods that explicitly build on or present large-scale pretrained/foundation video or world models.
}
\label{tab:sec5_taxonomy}
\begin{tabular*}{\textwidth}{@{\extracolsep{\fill}} p{1.9cm} p{4.55cm} c c c c c}
\toprule
Group & Method & Task-cond. & Action-cond. & Structure-aware & Foundation-scale & Main use \\
\midrule

\multirow{8}{*}{\parbox[t]{1.9cm}{Imagination-\\Based}}
& UniPi~\citep{unipi} & \cmark & -- & -- & \cmark & Plan. \\
& Video Language Planning~\citep{vlp} & \cmark & -- & -- & \cmark & Plan. \\
& Dreamitate~\citep{liang2024dreamitate} & -- & -- & -- & \cmark & Plan. \\
& RoboDreamer~\citep{zhou2024robodreamer} & \cmark & -- & -- & -- & Plan. \\
& ManipDreamer~\citep{li2025manipdreamer} & \cmark & -- & \cmark & -- & Plan. \\
& DreMa~\citep{barcellona2025dream} & -- & \cmark & \cmark & -- & Data \\
& PhysWorld~\citep{mao2025robot} & \cmark & -- & \cmark & -- & Plan. \\
& DreamGen~\citep{jang2025dreamgen} & \cmark & -- & -- & \cmark & Data \\
\midrule

\multirow{7}{*}{\parbox[t]{1.9cm}{Action-\\Controllable}}
& IRASim~\citep{Zhu_2025_ICCV} & -- & \cmark & -- & -- & Plan. \\
& RoboEnvision~\citep{yang2025roboenvision} & \cmark & -- & -- & -- & Plan. \\
& RoboMaster~\citep{fu2026learning} & -- & \cmark & \cmark & -- & Data \\
& Ctrl-World~\citep{guo2026ctrlworld} & -- & \cmark & \cmark & -- & Eval. \\
& EnerVerse-AC~\citep{jiang2025enerverseac} & -- & \cmark & \cmark & -- & Eval. \\
& Interactive World Simulator~\citep{interworld} & -- & \cmark & -- & -- & Sim. \\
& EVA~\citep{eva} & \cmark & -- & -- & -- & Eval. \\
\midrule

\multirow{3}{*}{\parbox[t]{1.9cm}{Structure-\\Aware}}
& Mask2IV~\citep{li2025mask2iv} & \cmark & -- & \cmark & -- & Data \\
& TesserAct~\citep{zhen2025tesseract} & \cmark & -- & \cmark & -- & Sup. \\
& RoboVIP~\citep{wang2026robovip} & \cmark & -- & \cmark & -- & Data \\
\midrule

\multirow{8}{*}{\parbox[t]{1.9cm}{Foundation\\Video WM}}
& Vid2World~\citep{huang2026vidworld} & -- & \cmark & -- & \cmark & Sim. \\
& Genie Envisioner~\citep{liao2026genie} & \cmark & \cmark & -- & \cmark & Sim. \\
& DreamDojo~\citep{gao2026dreamdojo} & -- & \cmark & -- & \cmark & Sim. \\
& WoW~\citep{chi2025wow} & -- & -- & -- & \cmark & Plan. \\
& UnifoLM-WMA-0~\citep{unifolm-wma-0} & \cmark & \cmark & -- & -- & Sim. \\
& Cosmos Predict 2.5~\citep{CosmosPredict} & \cmark & -- & -- & \cmark & Sim. \\
& GigaWorld-0~\citep{team2025gigaworld} & -- & -- & \cmark & \cmark & Data \\
& ABot-PhysWorld~\citep{chen2026abotphysworldinteractiveworldfoundation} & -- & \cmark & -- & \cmark & Sim. \\
\bottomrule
\end{tabular*}
\end{table*}

\subsection{Problem Setting and Scope}

An important route to embodied world modeling represents the future directly in image or video space. In this setting, the model predicts the visual evolution of robot--environment interaction from the current observation, task specification, and often a sequence of candidate actions. Unlike generic video synthesis, robotic video generation is subject to substantially stronger requirements: the predicted future should be not only visually plausible, but also temporally coherent, action-consistent, physically credible, and useful for downstream decision making. For this reason, robotic video generation should not be understood merely as a perceptual generation problem, but as a concrete mechanism for constructing visually explicit world models that support robotics policy learning, planning, evaluation, simulation, and data generation. Recent progress in large-scale video generation backbones, such as CogVideoX, has played an important enabling role by showing that long-horizon and high-fidelity spatiotemporal generation can be learned at scale and later adapted to embodied settings \citep{yangcogvideox}. We focus on video-based world models in this section because they form a rapidly growing and practically influential branch of recent work, not because pixel-level prediction is assumed to be the most compact or universally optimal representation for embodied control. 

In this survey, we also treat task or language conditioning as a form of high-level action. Under this view, robotic video generation includes not only low-level action-conditioned rollout, but also text- or task-guided visual prediction that specifies what future should be realized before low-level control is grounded. Accordingly, the key question is not simply whether a model can generate visually convincing future videos, but whether those futures are actionable: whether they remain faithful to the conditioning actions, preserve physically plausible interaction dynamics, and can be translated into executable robot behavior. In robotics, the value of a video world model depends on whether it preserves action consequences, interaction structure, and physical regularities in a way that improves policy behavior. Following this viewpoint, as shown in Fig.~\ref{fig:section5}, we organize the literature into four stages of progression: video generation as imagination for policy learning, action-controllable rollout models, structure-aware generation with richer interaction priors, and foundation-scale video backbones adapted into reusable world models. Table~\ref{tab:sec5_taxonomy} summarizes representative methods under this capability-oriented taxonomy.

\subsection{Video Generation as Imagination for Policy Learning}
A first class of methods uses video generation primarily as an imagination engine for policy learning. The central idea is to exploit strong generative priors to synthesize future task executions and then convert these imagined futures into supervision for robot control. In this line, the video model is valuable not because it produces visually impressive clips, but because it expands supervision beyond the narrow support of collected robot trajectories.

A closely related branch is text- or task-guided robotic video generation. In our definition, language can be viewed as a high-level action that specifies what future should be realized. From this perspective, methods such as UniPi~\citep{unipi} and Video Language Planning~\citep{vlp} are not merely performing perceptual synthesis, but robotic world modeling in which semantic actions are translated into predictive visual trajectories. This branch is mainly useful for data amplification and visual planning, as it provides task-relevant demonstrations and future-oriented supervision before low-level action grounding.

Dreamitate \citep{liang2024dreamitate} is an early and representative example. It fine-tunes a video diffusion model on task-specific human demonstrations and, at test time, uses synthesized executions in a novel scene directly as action-guiding visual plans for real-world robot control. RoboDreamer \citep{zhou2024robodreamer} extends this direction through compositional world modeling, where instructions are decomposed into reusable primitives and generation is conditioned on these structured components, improving generalization to unseen combinations of objects and actions. ManipDreamer \citep{li2025manipdreamer} further strengthens this line by introducing an action tree representation together with depth and semantic visual guidance, improving instruction following as well as temporal and physical consistency.

A related but more explicit perspective is to reinterpret imagination as a learnable digital twin. DreMa \citep{barcellona2025dream} combines Gaussian Splatting with a physics simulator to reconstruct an explicit and manipulable scene representation, allowing the model to generate additional demonstrations for imitation learning. PhysWorld \citep{mao2025robot} addresses the gap between photorealistic motion and physically executable behavior by reconstructing a physical world model from generated videos and grounding predicted motion into robot actions through object-centric residual reinforcement learning. These methods move beyond pure visual imagination and begin to connect generated futures to physically meaningful execution.

This imagination paradigm also scales naturally toward synthetic data generation and high-level planning. When future generation is conditioned on task or language descriptions, the resulting videos can serve as high-level demonstration surrogates or task-relevant synthetic supervision. They can also function as visual plans for long-horizon decision making~\citep{vlp,lvp}. DreamGen \citep{jang2025dreamgen} adapts strong video generators to a target embodiment, synthesizes neural trajectories, and recovers executable actions through latent action modeling or inverse dynamics. Its central message is that stronger video world models can be used not only to regularize policies, but also to produce synthetic experience that improves downstream generalization. Taken together, these works establish the first major role of robotic video generation, namely to serve as an imagination engine that broadens the supervisory and planning signals available for policy learning.

\subsection{Toward Action-Controllable Video World Models}
A second class of methods shifts the emphasis from imagined supervision to explicit controllability. Here the central question is no longer whether a model can produce plausible future videos, but whether the generated future follows the commanded action sequence with sufficient precision to support manipulation reasoning and downstream control. This shift is important because, in embodied settings, a visually convincing rollout is of limited value if it does not respond faithfully to action intervention.

IRASim \citep{Zhu_2025_ICCV} is representative of this transition. It formulates robot manipulation as a trajectory-to-video problem and introduces frame-level action conditioning within each transformer block to strengthen alignment between individual actions and corresponding future frames. RoboEnvision \citep{yang2025roboenvision} focuses on long-horizon multi-task manipulation and emphasizes the difficulty of preserving semantic and temporal consistency over extended task evolution. RoboMaster \citep{fu2026learning} addresses more complex robot-object interactions through collaborative trajectory control. By decomposing manipulation into multiple phases and modeling the coupled motion of the robot arm and the manipulated object, it improves faithfulness under rich contact dynamics. Ctrl-World \citep{guo2026ctrlworld} pushes controllability further toward policy-in-the-loop rollout. It combines joint multi-view prediction, frame-level action control, and memory-based long-horizon generation so that predicted futures can support both policy evaluation and targeted policy improvement. EnerVerse-AC \citep{jiang2025enerverseac} follows a related direction and formulates the world model as an action-conditional multi-view generator that can act both as a data engine and as an evaluator for robotic inference. Interactive World Simulator \citep{interworld} pushes this line from controllable generation toward genuinely interactive simulation, emphasizing high-frequency, long-horizon, and stable policy-conditioned interaction for closed-loop rollout, demonstration collection, and policy evaluation. EVA \citep{eva} complements this direction from the perspective of post-training alignment by targeting the executability gap between visually plausible rollouts and physically executable robot behavior, using inverse-dynamics rewards to align video world models with smooth, embodiment-consistent action sequences.

Collectively, these works mark a decisive conceptual shift. For robotic video world models, fidelity is increasingly measured not only by realism, but also by action faithfulness, controllable interaction, and usefulness for closed-loop decision making.

\subsection{Structure-Aware Generation with Interaction and Geometry Priors}

A closely related thread improves controllability by introducing richer intermediate structure for interaction. Instead of conditioning only on low-dimensional action sequences, these methods encode masks, geometry, viewpoints, or identity cues that better preserve contact relations and scene structure. The underlying intuition is that robotic video generation becomes substantially more useful when the model is required to preserve explicit interaction structure, rather than merely synthesize visually convincing motion.

Mask2IV \citep{li2025mask2iv} is illustrative of this idea. It adopts a two-stage design that first predicts interaction trajectories for the actor and the object, and then generates a video conditioned on these trajectories. This removes the need for dense user-provided masks while retaining flexible control over interaction outcomes. TesserAct \citep{zhen2025tesseract} pushes structure further by extending the representation space from 2D video to a 4D embodied world model over RGB, depth, and normal signals, improving spatial consistency and enabling stronger inverse dynamics and policy learning. RoboVIP \citep{wang2026robovip} focuses on a practical requirement of modern policy learners, namely temporally coherent multi-view observations. It introduces visual identity prompting to guide multi-view video diffusion and uses the resulting videos as scalable augmentation for manipulation data. 

This structure-aware view also connects robotic video world models to a broader line of structured and symbolic world modeling. While the methods above preserve structure inside generated visual futures, another family abstracts the world into predicates, object relations, affordances, or causal processes and predicts their transitions for planning~\citep{visualpredicator,pix2pred,exopredicator}. These approaches do not aim to improve visual fidelity; instead, they target more compact and compositional predictive variables that may be better aligned with long-horizon reasoning and executable control.

Although these methods differ in representation and objective, they share a common principle: richer structural priors can make generated futures more controllable, more consistent across views and contacts, and ultimately more useful for downstream embodied learning.

\subsection{From Video Backbones to Foundation World Models}
The most recent works reinterpret robotic video world models as general-purpose interactive predictors built by adapting large-scale video backbones. In this regime, video generation is no longer merely a downstream augmentation tool. It becomes a reusable substrate for simulation, planning, evaluation, and large-scale robot data production.

Vid2World \citep{huang2026vidworld} is a canonical example of this shift. Rather than training a robotic world model from scratch, it systematically transforms a pretrained video diffusion model into an interactive world model suitable for action-conditioned rollout. Genie Envisioner \citep{liao2026genie} extends this idea into a unified world foundation platform that integrates video world modeling with action decoding for robotic manipulation. DreamDojo \citep{gao2026dreamdojo} pushes the foundation-model regime further by pretraining on large-scale human egocentric videos, introducing continuous latent actions to bridge unlabeled human interaction and robot control, and then post-training the model for target embodiments. It shows that such a model can support long-horizon real-time rollout, policy evaluation, and model-based planning.

A complementary argument is made by WoW \citep{chi2025wow}. It emphasizes that physical intuition cannot be acquired from passive video observation alone, and instead trains a large generative world model on extensive robot interaction trajectories. By coupling generative rollout with inverse dynamics and critique, it explicitly closes the imagination-to-action loop. At the platform level, UnifoLM-WMA-0 \citep{unifolm-wma-0} and Cosmos Predict 2.5 \citep{CosmosPredict} further reflect the trend toward reusable world backbones, while GigaWorld-0 \citep{team2025gigaworld} makes the data-engine perspective fully explicit by combining a controllable video branch with a physically grounded 3D branch for large-scale embodied data synthesis.  ABot-PhysWorld \citep{chen2026abotphysworldinteractiveworldfoundation} extends this trend toward physics-aligned world foundation models. It explicitly targets physically plausible and action-controllable manipulation video generation.

Taken together, these works indicate a broader transition in the field. Robotic video generation is increasingly treated not as an isolated generative task, but as a foundation layer for interactive world modeling.

\subsection{Technical Progression and Open Challenges}

Viewed together, the literature reveals a clear technical progression. Early methods such as Dreamitate, RoboDreamer, DreMa, ManipDreamer, DreamGen, and PhysWorld mainly treat video generation as imagination that supplies additional supervision for policy learning \citep{liang2024dreamitate,zhou2024robodreamer,barcellona2025dream,li2025manipdreamer,jang2025dreamgen,mao2025robot}. The next wave, including IRASim, RoboEnvision, RoboMaster, Ctrl-World, EnerVerse-AC, Interactive World Simulator, and EVA, makes action alignment, controllable rollout, interactive usability, executability, and evaluation utility central objectives \citep{Zhu_2025_ICCV,yang2025roboenvision,fu2026learning,guo2026ctrlworld,jiang2025enerverseac,interworld,eva}. A parallel line introduces richer interaction structure through masks, geometry, and multi-view identity cues, as in Mask2IV, TesserAct, and RoboVIP \citep{li2025mask2iv,zhen2025tesseract,wang2026robovip}. The newest systems, including Vid2World, Genie Envisioner, DreamDojo, WoW, ABot-PhysWorld, UnifoLM-WMA-0, Cosmos Predict 2.5, and GigaWorld-0, increasingly elevate robotic video generation into a reusable foundation layer for embodied world modeling \citep{huang2026vidworld,liao2026genie,gao2026dreamdojo,chi2025wow,chen2026abotphysworldinteractiveworldfoundation,unifolm-wma-0,CosmosPredict,team2025gigaworld}.

This progression also clarifies the central bottleneck of the field. The key challenge is no longer simply to generate realistic futures. It is to generate futures that remain causally aligned with robot actions, physically and kinematically self-consistent over long horizons, coherent across views and embodiments, stable under interaction, and executable enough to support real policy improvement. For robotics, therefore, the true value of video generation lies in turning future prediction into a controllable, interactive, and actionable interface between perception and decision making.
\section{World Model for Other Applications}
\label{sec:other_app}
%\hbh{drafted -3/25}
\subsection{World Model for Navigation}

World models have become a useful abstraction for embodied navigation, where agents must act under severe partial observability and reason about spaces, objects, or paths that are not yet visible. Instead of treating navigation as a purely reactive next-step decision problem, this line of work uses world models to imagine action-conditioned future observations, construct future-aware planning states, or derive value-like signals from imagined trajectories before the agent physically moves \citep{pathdreamer,nwm,vistav2,wmnav,xvwm}. In this sense, the world model turns unseen space into a predictive planning substrate for reasoning about future visibility, traversability, and goal progress \citep{pathdreamer,nwm}.

Early works emphasize look-ahead prediction of unseen observations. Pathdreamer generates plausible future 360$^\circ$ RGB, depth, and semantic observations for unvisited indoor viewpoints, showing that planning with imagined observations can substantially reduce the gap to planning with true future observations \citep{pathdreamer}. VISTA introduces an ``imagine-and-align'' strategy for instruction-conditioned visual imagination \citep{vista}, while VISTAv2 extends this into a navigation world model that rolls out egocentric futures under candidate actions and projects them into an online value map for planning \citep{vistav2}. A parallel trend scales this idea into controllable video predictors: NWM formulates controllable video generation explicitly as a navigation world model \citep{nwm}; SparseVideoNav replaces dense long-horizon rollout with sparse future generation for faster deployment \citep{sparsevideonav}; and EgoWM adapts pretrained Internet-scale video diffusion models into action-conditioned egocentric world models through lightweight conditioning \citep{egowm}. Taken together, these methods show that the value of world models in navigation lies less in visual realism itself than in exposing hidden future structure in a form usable for planning, trajectory ranking, and value estimation.
\subsection{World Model for Autonomous Driving}

World models have likewise become an important paradigm in autonomous driving, where they increasingly unify perception, prediction, planning, and simulation. Compared with robotic manipulation, driving places stronger demands on long-horizon forecasting, multi-agent interaction, structured geometry, and safety-critical planning. Accordingly, driving world models often learn future-evolving scene representations---in image space, multi-view space, occupancy space, or compact latent space---and use them to support downstream planning or end-to-end driving decisions \citep{mile,gaia,occworld,drivewm}. Early and representative works already reveal two complementary routes. One emphasizes compact or structured predictive states: MILE learns a latent dynamics model with geometric inductive bias for urban driving \citep{mile}, while OccWorld formulates world modeling in 3D occupancy space so that ego motion and scene evolution are represented in a planning-compatible form \citep{occworld}. The other makes the generative world-model view more explicit: GAIA-1 casts driving world modeling as multimodal sequence modeling over video, text, and action tokens \citep{gaia}, and DriveDreamer uses diffusion-based modeling with structural constraints to capture complex traffic evolution from real driving data \citep{drivedreamer}.

More recent work pushes these directions toward planning-oriented and unified driving intelligence. Drive-WM generates controllable multiview future videos and uses imagined multi-future rollouts together with image-based rewards to select safer trajectories \citep{drivewm}. UniDWM further argues for a structure- and dynamics-aware latent world representation as a unified substrate for perception, prediction, and planning \citep{unidwm}. DriveWorld-VLA strengthens the connection between world modeling and action generation by using latent world states as the planner's decision state, allowing action-conditioned imagination to guide control without expensive pixel rollout \citep{dwvla}. From the scaling perspective, DriveVLA-W0 shows that future-image prediction through world modeling provides dense self-supervision that improves end-to-end driving VLAs beyond low-dimensional action supervision alone \citep{drivevlaw}. Furthering this hierarchical synergy, SteerVLA~\citep{steervla26} conceptually frames the high-level VLM as a semantic world model that generates fine-grained, common-sense reasoning to steer a low-level VLA policy through complex, long-tail driving maneuvers. Overall, world models in autonomous driving are best understood as a bridge from passive scene understanding to predictive driving intelligence \citep{gaia,drivewm,unidwm}: some methods use them as controllable simulators of future scenes, some as structured state spaces for planning, and some as dense predictive supervision for scaling end-to-end driving policies \citep{occworld,drivevlaw}. Across these variants, the shared intuition is the same: safe driving requires not only recognizing the current scene, but reasoning over how it may evolve under ego behavior and surrounding traffic dynamics \citep{mile,drivedreamer,unidwm}.
\section{Benchmarks, Datasets, and Results}
\label{sec:benchmark}
\todo[inline, color=myyellow]{Assigned to: Tuo An}
%\ant{drafted by ant 03/01} \ant{revised by 3/11} %\ant{revised by 3/12}
%\hbh{table-revised}
%\ant{4/2 first completed version, to be compressed}

\subsection{Benchmarks for World Model Evaluation}
In embodied intelligence, evaluating world models differs fundamentally from evaluating video generation models in conventional computer vision. In robotics, the value of a world model depends on whether it can generate action-conditioned future states that remain consistent with real physical dynamics. This requires the model to capture robot--environment interactions beyond surface-level realism, functioning instead as a faithful predictor of physically plausible and temporally coherent future observations.

More importantly, such a model should respond reliably to action interventions, remain coherent over long horizons, and support downstream tasks such as policy learning, planning, and evaluation. From this policy-centric perspective, visual realism alone is neither necessary nor sufficient~\citep{shang2026worldarena}, since rollouts may look convincing while still violating dynamics in ways that break closed-loop control~\citep{qin2025worldsimbench, li2025worldeval}. Accordingly, we organize existing embodied world model benchmarks into three complementary categories: (i) action-conditioned generation and open-loop predictive quality, (ii) closed-loop task utility and policy evaluation, and (iii) physical consistency, controllability, and executability diagnostics.
\subsubsection{Action-conditioned generation and open-loop predictive quality}
%\ant{4/3 compressed}

The first dimension evaluates embodied world models in an open-loop setting. Given the current observation together with an action sequence, language instruction, or task specification, the model is asked to autoregressively generate future observations without being embedded into a planner or control loop. The key question is whether the predicted future remains faithful to the commanded behavior over time, in terms of semantic correctness, temporal coherence, and action responsiveness, rather than visual plausibility alone. Open-loop benchmarks are attractive because they are relatively easy to scale and standardize, although their results should be interpreted with care.

Recent benchmarks in this direction have become increasingly embodied. Rather than treating robot video generation as generic video synthesis, RBench~\citep{deng2026rethinking} and EWMBench~\citep{yue2025ewmbench} evaluate whether generated futures preserve the task-relevant structure of embodied interaction. RBench emphasizes structural consistency, physical plausibility, and action completeness across diverse robotic tasks and embodiments. EWMBench adopts a more factorized view, separating scene consistency, motion correctness, and semantic alignment. Together, they reflect a broader shift in open-loop evaluation from appearance-level realism to interaction-faithful prediction.

Other benchmarks further connect open-loop prediction to downstream utility. DreamGen Bench~\citep{jang2025dreamgen} evaluates instruction following and physics alignment, asking whether generated rollouts are useful as synthetic experience for policy learning rather than merely realistic. EVA-Bench~\citep{chi2025empowering} complements this view by emphasizing long-horizon anticipation and out-of-domain robustness under variations in viewpoint, scene layout, and motion distribution. Overall, these benchmarks suggest that strong open-loop world models must do more than generate plausible futures: they must remain action-grounded, physically sensible, and robust enough to support downstream embodied use.

\subsubsection{Closed-loop task utility and policy evaluation}
%\ant{4/3 compressed}

While open-loop benchmarks evaluate whether a world model can generate action-conditioned futures, closed-loop benchmarks ask whether those predictions remain useful inside an interactive decision loop. In this setting, the world model is evaluated not as a passive predictor, but as an environment simulator, policy evaluator, or planning substrate that directly influences action selection over time. The focus therefore shifts from predictive plausibility to decision utility: whether the model preserves the task-relevant dynamics needed for policy ranking, value estimation, planning, and ultimately task success. This makes closed-loop evaluation more aligned with embodied intelligence, since small modeling errors can accumulate and break control once the agent repeatedly acts on model-generated futures.

Recent benchmarks in this category differ in protocol, but share the same principle: a useful embodied world model must support downstream decision making, not just realistic generation. WorldArena~\citep{shang2026worldarena} makes this explicit by evaluating world models not only with perceptual criteria, but also through functional roles such as synthetic data generation, policy evaluation, and action planning, highlighting the gap between visual realism and embodied utility. WorldEval~\citep{li2025worldeval} operationalizes this idea through comparative policy assessment, asking whether rollouts in a learned world model preserve the relative ordering of robot policies and checkpoints. WorldGym~\citep{quevedo2025worldgym} extends this setting by treating the learned model as an interactive environment for Monte Carlo evaluation, focusing on whether estimated policy values and success trends match those in the real world. Across these works, rank consistency, value fidelity, and decision reliability emerge as more informative criteria than pixel-level accuracy.

A stricter version of this evaluation places the world model directly inside a closed-loop planning pipeline and measures embodied task success. World-in-World~\citep{zhang2025world} is representative of this setting: by providing a unified interface for integrating heterogeneous world models into online planning tasks, it tests whether the model can improve control under iterative prediction and replanning. This is a harder setting than open-loop rollout evaluation because it exposes compounding errors that arise when prediction and action interact over time. Overall, recent evidence suggests that visual plausibility is only a weak proxy for control utility, whereas action-grounded consistency and controllability are much more reliable indicators of downstream embodied usefulness.

\subsubsection{Physical consistency, controllability, and executability diagnostics}
%\ant{4/3 compressed}

While open-loop benchmarks evaluate predictive quality and closed-loop benchmarks evaluate downstream utility, diagnostic benchmarks ask a more targeted question: which properties of a generated rollout determine whether it is actually usable for embodied control? This dimension focuses on whether predicted futures preserve the physical and action-relevant structure required for execution, including consistency with dynamics, responsiveness to action interventions, and recoverability into valid control signals. Rather than measuring overall prediction quality or end-task success, it probes the specific failure modes that often explain why visually plausible rollouts still fail in planning, policy evaluation, or execution.

WorldSimBench~\citep{qin2025worldsimbench} is representative of this direction. It combines perceptual evaluation with manipulative evaluation, asking not only whether generated videos look realistic, but also whether they remain sufficiently consistent with action and environment dynamics to support inverse-dynamics recovery and downstream control. WoW-World-Eval~\citep{fan2026wow} provides a broader but closely related perspective. Although it spans perception, planning, prediction, execution, and generalization, it is especially relevant here because it introduces physical-law and execution-oriented criteria, including an IDM-based Turing Test for whether generated videos induce plausible and executable actions. Together, these benchmarks make clear that visual plausibility alone is insufficient: generated rollouts must also preserve physically grounded and operationally executable action consequences.

Related evidence appears in adjacent domains such as autonomous driving. DrivingGen~\citep{zhou2026drivinggen} evaluates generative driving world models not only by visual realism, but also by trajectory plausibility, temporal coherence, and controllability under ego conditioning. Its results reveal a trade-off between appearance quality and physically reliable motion generation, reinforcing the broader point that action-conditioned world models should be judged by control-relevant dynamics rather than visual appeal alone.

A complementary diagnostic direction examines the component abilities underlying world modeling itself. WM-ABench~\citep{gao2025vision} fits this role by decomposing evaluation into atomic capabilities such as spatial and temporal understanding, motion perception, mechanistic simulation, and controlled counterfactual reasoning. Although such benchmarks do not directly test rollout executability, they are useful for identifying which internal predictive or causal capacities are missing when a model fails in more integrated open-loop or closed-loop settings.

Overall, these three categories form a layered evaluation framework for embodied world models. Open-loop benchmarks test whether the model can generate coherent action-conditioned futures; closed-loop benchmarks test whether those predictions remain useful for planning and policy evaluation; diagnostic benchmarks test whether the generated futures are physically grounded, controllable, and executable. Together, they highlight a broader lesson from recent work: no single metric is sufficient for embodied world model evaluation. A strong model must not only predict plausible futures, but also preserve the action-relevant structure needed for reliable control.

\begin{table}[t]
\centering
\footnotesize
\caption{Core attributes of representative datasets/resources for embodied world model training. Benchmark-only resources discussed in the evaluation section are excluded.}
\label{tab:wm_datasets_attr_full}
%\begin{threeparttable}
\renewcommand{\arraystretch}{1.08}
\begin{tabular}{l c c c c c c c}
\toprule
\textbf{Name} & \textbf{Year} & \textbf{Source} & \textbf{X-Emb.} & \textbf{Act.} & \textbf{Obs./3D} & \textbf{Lang.} & \textbf{M/C} \\
\midrule
RoVid-X~\citep{deng2026rethinking}& 2026 & Real/Robot video & \pmark & \pmark & \pmark & \cmark & \xmark \\
\makecell[l]{Open X-Embodiment (OXE) \\ \citep{o2024open}} & 2024 & Real & \cmark & \cmark & \pmark & \pmark & \pmark \\
DROID~\citep{khazatsky2024droid} & 2024 & Real & \xmark & \cmark & \pmark & \cmark & \pmark \\
BridgeData V2~\citep{walke2023bridgedata} & 2023 & Real & \xmark & \cmark & \pmark & \cmark & \pmark \\
AgiBot World~\citep{bu2025agibot} & 2025 & Real & \pmark & \cmark & \pmark & \cmark & \pmark \\
\makecell[l]{Galaxea Open-World Datase\\~\citep{jiang2025galaxea}} & 2025 & Real & \pmark & \cmark & \pmark & \cmark & \pmark \\
Humanoid Everyday~\citep{zhao2025humanoid} & 2025 & Real & \pmark & \cmark & \cmark & \cmark & \cmark \\
RoboMIND 2.0~\citep{hou2025robomind} & 2025 & Real+Sim & \cmark & \cmark & \pmark & \cmark & \cmark \\
FastUMI-100K~\citep{liu2025fastumi} & 2025 & Real & \pmark & \cmark & \cmark & \cmark & \pmark \\
BRMData~\citep{zhang2024empowering} & 2024 & Real & \pmark & \cmark & \cmark & \pmark & \pmark \\
UMI~\citep{chi2024universal} & 2024 & Real & \pmark & \cmark & \pmark & \pmark & \xmark \\
MV-UMI~\citep{rayyan2025mv} & 2025 & Real & \cmark & \cmark & \cmark & \pmark & \xmark \\
ActiveUMI~\citep{zeng2025activeumi} & 2025 & Real & \pmark & \cmark & \cmark & \pmark & \xmark \\
TWIST2~\citep{ze2025twist2} & 2025 & Real & \pmark & \cmark & \pmark & \xmark & \xmark \\
DexWild~\citep{taodexwild} & 2025 & Human+Robot & \cmark & \pmark & \pmark & \xmark & \xmark \\
EgoMimic~\citep{kareer2025egomimic} & 2025 & Human+Robot & \pmark & \pmark & \cmark & \xmark & \xmark \\
PHSD / In-N-On~\citep{cai2025n} & 2025 & Human ego & \xmark & \pmark & \pmark & \xmark & \xmark \\
UniHand~\citep{luo2025being} & 2025 & Human video & \pmark & \pmark & \pmark & \cmark & \xmark \\
UniHand 2.0~\citep{luo2026being} & 2026 & Human+Robot+VLM & \cmark & \cmark & \pmark & \cmark & \xmark \\
Hoi!~\citep{engelbracht2025hoi} & 2025 & Human+Robot & \cmark & \cmark & \cmark & \xmark & \cmark \\
FreeTacMan~\citep{wu2025freetacman} & 2025 & Robot-free & \pmark & \cmark & \cmark & \xmark & \cmark \\
\makecell[l]{Humanoid Visual-Tactile-Action\\~\citep{kwon2025humanoid}} & 2025 & Real & \xmark & \cmark & \pmark & \xmark & \cmark \\
VTDexManip~\citep{liu2025vtdexmanip} & 2025 & Human tactile & \xmark & \pmark & \pmark & \xmark & \cmark \\
RH20T~\citep{fang2023rh20t} & 2023 & Real & \pmark & \cmark & \pmark & \cmark & \cmark \\
RH20T-P~\citep{chen2025rh20t} & 2025 & Real & \pmark & \cmark & \pmark & \cmark & \cmark \\
RoboTwin 2.0~\citep{chen2025robotwin} & 2025 & Sim & \cmark & \cmark & \pmark & \cmark & \pmark \\
Action100M~\citep{chen2026action100m} & 2026 & Web video & \xmark & \pmark & \xmark & \cmark & \xmark \\
\bottomrule
\end{tabular}

\begin{tablenotes}[flushleft]
\footnotesize
\item \textbf{X-Emb.}: cross-embodiment coverage. \textbf{Act.}: explicit action supervision or aligned action proxy. \textbf{Obs./3D}: strong observation support beyond basic monocular RGB, e.g., multi-view, depth, LiDAR, or 3D annotations. \textbf{Lang.}: language/task conditioning. \textbf{M/C}: multimodal or contact-rich signals such as force, tactile, audio, or dense proprioceptive/contact cues.
\item \cmark\ denotes strong support, \pmark\ denotes partial/moderate support, and \xmark\ denotes absent or not emphasized.
\end{tablenotes}
\vspace{-1em}
%\end{threeparttable}
\end{table}

\vspace{-2em}
\subsection{Datasets for World Model Training}
%\ant{drafted by 3/17} \ant{revised by 3/19, table draft, just a rough version}

% 感觉很难显式地对现有 dataset 做出分类
% Datasets are a central determinant of what embodied world models can learn, because they define the action-conditioned state transitions, interaction patterns, and task structures available for supervision. In contrast to passive visual corpora, embodied world model training requires data that reflects how the world changes under intervention, often coupling perceptual observations with actions, proprioceptive signals, embodiment constraints, and task-relevant outcomes. From this perspective, existing datasets differ not only in size, modality, or domain coverage, but more fundamentally in the kind of predictive structure they provide: broad robot trajectory corpora support general dynamics learning, interaction-rich datasets better expose decision-relevant and long-horizon consequences, and large heterogeneous corpora facilitate scaling across embodiments and modalities. We therefore group datasets in this section by the training role they play for embodied world models, which more directly reflects their relevance than a simple catalog by source or platform.
% \subsubsection{General-purpose robot trajectory datasets}

% \subsubsection{Interaction-rich and decision-relevant datasets}
% \subsubsection{Cross-embodiment and multimodal scaling datasets}
% 尝试使用多维表格的方式：
Complementary to benchmarks, which specify how embodied world models should be evaluated, training datasets determine what kinds of experience such models can learn from in the first place. For embodied intelligence, such data are not merely collections of videos, but samples of \textit{agent--environment transitions} that may couple observations with actions, task progression, embodiment-specific constraints, and physical interaction dynamics. As a result, the value of a dataset is not defined by scale alone, but by whether it provides sufficiently rich action-conditioned transitions, long-horizon task structure, diversity across scenes and embodiments, and coverage of manipulation-relevant physical signals. These properties jointly determine whether a world model can acquire dynamics priors that are genuinely useful for prediction, planning, and control.
\begin{table}[t]
\centering
\footnotesize
\caption{Relevance of representative datasets/resources to embodied world-modeling capabilities.}
\label{tab:wm_datasets_relevance_full}
\begin{threeparttable}
\renewcommand{\arraystretch}{1.06}
\setlength{\tabcolsep}{3.8pt}
\begin{tabular*}{\textwidth}{@{\extracolsep{\fill}} p{5cm} c c c c c c}
\toprule
\textbf{Name} &
\makecell[c]{\textbf{General}\\\textbf{Traj.}} &
\makecell[c]{\textbf{Long-}\\\textbf{Horizon}} &
\makecell[c]{\textbf{X-Emb.}\\\textbf{Scaling}} &
\makecell[c]{\textbf{Human}\\\textbf{Prior}} &
\makecell[c]{\textbf{Contact}\\\textbf{/ Physics}} &
\makecell[c]{\textbf{Synth.}\\\textbf{/ Recipe}} \\
\midrule
RoVid-X~\citep{deng2026rethinking} & \pmark & \pmark & \pmark & \xmark & \pmark & \xmark \\
\makecell[l]{Open X-Embodiment (OXE)\\~\citep{o2024open}} & \cmark & \pmark & \cmark & \xmark & \pmark & \xmark \\
DROID~\citep{khazatsky2024droid} & \cmark & \pmark & \xmark & \xmark & \pmark & \xmark \\
\makecell[l]{BridgeData V2\\~\citep{walke2023bridgedata}} & \cmark & \pmark & \xmark & \xmark & \pmark & \xmark \\
AgiBot World~\citep{bu2025agibot} & \cmark & \pmark & \pmark & \xmark & \pmark & \xmark \\
\makecell[l]{Galaxea Open-World Dataset\\~\citep{jiang2025galaxea}} & \cmark & \pmark & \pmark & \xmark & \pmark & \xmark \\
\makecell[l]{Humanoid Everyday\\~\citep{zhao2025humanoid}} & \cmark & \cmark & \pmark & \xmark & \cmark & \xmark \\
RoboMIND 2.0~\citep{hou2025robomind} & \cmark & \cmark & \cmark & \xmark & \cmark & \pmark \\
FastUMI-100K~\citep{liu2025fastumi} & \cmark & \cmark & \pmark & \xmark & \pmark & \xmark \\
BRMData~\citep{zhang2024empowering} & \cmark & \cmark & \pmark & \xmark & \pmark & \xmark \\
UMI~\citep{chi2024universal} & \pmark & \pmark & \pmark & \cmark & \xmark & \xmark \\
MV-UMI~\citep{rayyan2025mv} & \pmark & \pmark & \cmark & \cmark & \xmark & \xmark \\
ActiveUMI~\citep{zeng2025activeumi} & \pmark & \pmark & \pmark & \cmark & \xmark & \xmark \\
TWIST2~\citep{ze2025twist2} & \pmark & \pmark & \pmark & \xmark & \pmark & \xmark \\
DexWild~\citep{taodexwild} & \pmark & \pmark & \cmark & \cmark & \xmark & \xmark \\
EgoMimic~\citep{kareer2025egomimic} & \xmark & \pmark & \pmark & \cmark & \xmark & \xmark \\
PHSD / In-N-On\\~\citep{cai2025n} & \xmark & \pmark & \xmark & \cmark & \xmark & \xmark \\
UniHand~\citep{luo2025being} & \xmark & \pmark & \pmark & \cmark & \xmark & \xmark \\
UniHand 2.0~\citep{luo2026being} & \pmark & \pmark & \cmark & \cmark & \xmark & \cmark \\
Hoi!~\citep{engelbracht2025hoi} & \pmark & \pmark & \cmark & \pmark & \cmark & \xmark \\
FreeTacMan~\citep{wu2025freetacman} & \pmark & \pmark & \pmark & \pmark & \cmark & \xmark \\
\makecell[l]{Humanoid Visual-Tactile-Action\\~\citep{kwon2025humanoid}} & \pmark & \pmark & \xmark & \xmark & \cmark & \xmark \\
VTDexManip~\citep{liu2025vtdexmanip} & \xmark & \pmark & \xmark & \cmark & \cmark & \xmark \\
RH20T~\citep{fang2023rh20t} & \pmark & \pmark & \pmark & \xmark & \cmark & \xmark \\
RH20T-P~\citep{chen2025rh20t} & \pmark & \pmark & \pmark & \xmark & \cmark & \xmark \\
\makecell[l]{RoboTwin 2.0\\~\citep{chen2025robotwin}} & \pmark & \pmark & \cmark & \xmark & \pmark & \cmark \\
Action100M~\citep{chen2026action100m} & \xmark & \pmark & \xmark & \pmark & \xmark & \cmark \\
\bottomrule
\end{tabular*}

\begin{tablenotes}[flushleft]
\footnotesize
\item \cmark\ strong relevance; \pmark\ partial relevance; \xmark\ limited or no direct relevance.
\end{tablenotes}
\end{threeparttable}
\end{table}

Existing resources relevant to embodied world model training are rarely well characterized by a single taxonomy. A given dataset may simultaneously serve as a general-purpose trajectory corpus, a cross-embodiment aggregation resource, a human-to-robot prior, and a multimodal interaction dataset. For this reason, rather than forcing datasets into mutually exclusive categories, we compare them along several complementary dimensions. Table~\ref{tab:wm_datasets_attr_full} summarizes their core data attributes, including embodiment coverage, action supervision, observation and 3D support, language conditioning, and multimodal or contact-rich signals. Table~\ref{tab:wm_datasets_relevance_full} further organizes the same resources by the kinds of world-modeling capability they are most likely to support, including general trajectory pretraining, long-horizon modeling, cross-embodiment scaling, human-prior transfer, contact- and physics-aware modeling, and synthetic or recipe-driven data scaling.

Taken together, these comparisons suggest that current training resources are better understood as spanning several parallel axes rather than falling into disjoint groups. Large-scale robot trajectory corpora provide the basic transition coverage needed for action-conditioned prediction, while cross-embodiment datasets encourage more transferable dynamics priors across platforms. Human-video and human-to-robot resources offer an additional route for learning interaction regularities beyond robot-collected trajectories, whereas tactile-, force-, and contact-rich datasets are particularly important for grounding executability and physical consistency. In parallel, synthetic data and aggregated data recipes broaden the controllable variation available for training. This multi-axis view also clarifies a central limitation of the current landscape: despite the rapid growth of available resources, failure recovery, decision-sensitive variation, and dense physically grounded supervision remain much scarcer than large-scale successful demonstrations.
\begin{table}[!htbp]
\centering
\footnotesize
\caption{Representative results on the LIBERO standard 4-suite benchmark, grouped by how world modeling is integrated with policy learning. Methods with no directly reported number under the standard Spatial/Object/Goal/Long protocol are omitted.}
\label{tab:all_libero_total}
\begin{threeparttable}
\renewcommand{\arraystretch}{1.08}
\setlength{\tabcolsep}{5.0pt}
\begin{tabular}{l l c c c c c}
\toprule
\textbf{Group} & \textbf{Method} & \textbf{Spatial} & \textbf{Object} & \textbf{Goal} & \textbf{Long} & \textbf{Avg} \\
\midrule
\multirow{3}{*}{Decoupled}
& UniPi~\citep{unipi} & -- & -- & -- & 0.0 & -- \\
& MimicVideo~\citep{mimicvideo} & 94.2 & 96.8 & 90.6 & 94.0 & 93.9 \\
& Say-Dream-ACT~\citep{saydream} & 99.4 & 99.2 & 98.6 & 95.4 & 98.1 \\
\midrule
\multirow{4}{*}{Single-backbone}
& UVA~\citep{UVA} & -- & -- & -- & 90.0 & -- \\
& VideoPolicy~\citep{VideoPolicy} & -- & -- & -- & 94.0 & -- \\
& Cosmos Policy~\citep{CosmosPolicy} & 98.1 & 100.0 & 98.2 & 97.6 & 98.5 \\
& UD-VLA~\citep{udvla} & 94.1 & 95.7 & 91.2 & 89.6 & 92.7 \\
\midrule
\multirow{2}{*}{MoE / MoT}
& Motus~\citep{motus} & 96.8 & 99.8 & 96.6 & 97.6 & 97.7 \\
& LingBot-VA~\citep{lingbotva} & 98.5 & 99.6 & 97.2 & 98.5 & 98.5 \\
\midrule
\multirow{8}{*}{Unified VLA}
& RynnVLA-002~\citep{rynnvla002} & 99.0 & 99.8 & 96.4 & 94.4 & 97.4 \\
& DreamVLA~\citep{dreamvla} & 97.5 & 94.0 & 89.5 & 89.5 & 92.6 \\
& UniVLA~\citep{bu2025univla} & 96.5 & 96.8 & 95.6 & 92.0 & 95.2 \\
& Unified VLA~\citep{unifiedvla} & 95.4 & 98.8 & 93.6 & 94.0 & 95.5 \\
& CoWVLA~\citep{cowvla} & 97.2 & 97.8 & 94.6 & 92.8 & 95.6 \\
& F1~\citep{f1vla} & 98.2 & 97.8 & 95.4 & 91.3 & 95.7 \\
%& InternVLA-M1 & 98.0 & 99.0 & 93.8 & 92.6 & 95.9 \\
& TriVLA~\citep{TriVLA} & 91.2 & 93.8 & 89.8 & 73.2 & 87.0 \\
\midrule
\multirow{2}{*}{Latent-space WM}
& VLA-JEPA~\citep{vlajepa} & 96.2 & 99.6 & 97.2 & 95.8 & 97.2 \\
& JEPA-VLA~\citep{jepavla} & 97.2 & 98.0 & 95.6 & 94.8 & 96.4 \\
\bottomrule
\end{tabular}

\begin{tablenotes}[flushleft]
\footnotesize
\item Methods are grouped by the manner in which world modeling is incorporated into policy learning, rather than by publication year alone.
\item \textbf{Avg} denotes the average success rate over the four LIBERO suites when directly reported by the original paper. ``--'' indicates that the corresponding suite-level result was not directly reported under the standard Spatial/Object/Goal/Long protocol.
\end{tablenotes}
\end{threeparttable}
\vspace{-1em}

\end{table}

% \subsubsection{Downstream utility}
\subsection{Representative Results on Common Benchmarks}
%\ant{drafted by 3/30} \ant{4/3; compressed version}

\begin{table}[t]
\centering
\footnotesize
\caption{Representative results on RoboTwin, CALVIN, and SIMPLER-style benchmarks, grouped by how world modeling is integrated with policy learning. Methods with no directly reported number under the corresponding protocols are omitted.}
\label{tab:all_other_total}
\begin{threeparttable}
\renewcommand{\arraystretch}{1.08}
\setlength{\tabcolsep}{4.8pt}
\begin{tabular}{l l c c c c c c c}
\toprule
\textbf{Group} & \textbf{Method} & \textbf{RT-A} & \textbf{RT-B} & \textbf{C-A} & \textbf{C-D} & \textbf{S-G} & \textbf{S-W} & \textbf{S-O} \\
\midrule
\multirow{6}{*}{Decoupled}
& UniPi~\citep{unipi} & -- & -- & 0.92 & -- & -- & -- & -- \\
& VidMan~\citep{vidman} & -- & -- & 3.42 & -- & -- & -- & -- \\
& Vidar~\citep{vidar} & 65.8 & 17.5 & -- & -- & -- & -- & -- \\
& VPP~\citep{vpp} & -- & -- & 4.33 & -- & -- & -- & -- \\
& Video2Act~\citep{video2act} & 54.6 & 54.1 & -- & -- & -- & -- & -- \\
& MimicVideo~\citep{mimicvideo} & -- & -- & -- & -- & -- & -- & 46.9/56.3 \\
\midrule
\multirow{2}{*}{Single-backbone}
& VideoVLA~\citep{videovla} & -- & -- & -- & -- & 73.1/62.8 & 53.1 & 63.0 \\
& UD-VLA~\citep{udvla} & -- & -- & -- & 4.64 & -- & 62.5 & -- \\
\midrule
\multirow{6}{*}{MoE / MoT}
& Motus~\citep{motus} & 88.7 & 87.0 & -- & -- & -- & -- & -- \\
& LingBot-VA~\citep{lingbotva} & 92.9 & 91.6 & -- & -- & -- & -- & -- \\
& LingBot-VLA~\citep{wu2026pragmatic} & 88.6 & 86.7 & -- & -- & -- & -- & -- \\
& BagelVLA~\citep{bagelvla} & 75.3 & 20.9 & 4.41 & -- & -- & -- & -- \\
& FRAPPE~\citep{frappe} & 57.5 & 25.5 & -- & -- & -- & -- & -- \\
\midrule
\multirow{8}{*}{Unified VLA}
& GR-1~\citep{GR-1} & -- & -- & 3.06 & 4.21 & -- & -- & -- \\
& UP-VLA~\citep{upvla} & -- & -- & 4.08 & 4.42 & -- & -- & -- \\
& DreamVLA~\citep{dreamvla} & -- & -- & 4.44 & -- & -- & -- & -- \\
& Unified VLA~\citep{unifiedvla} & -- & -- & 4.41 & 4.63 & -- & 69.8 & -- \\
& CoWVLA~\citep{cowvla} & -- & -- & 4.21 & 4.47 & 60.9 & 76.0 & -- \\
& F1~\citep{f1vla} & -- & -- & -- & -- & -- & -- & 72.9 \\
& InternVLA-A1~\citep{cai2026internvla} & 89.4 & 87.0 & -- & -- & -- & -- & -- \\
& HALO~\citep{halo} & 80.5 & 26.4 & -- & -- & -- & -- & -- \\
& TriVLA~\citep{TriVLA} & -- & -- & 4.37 & -- & -- & -- & -- \\
\midrule
\multirow{3}{*}{Latent-space WM}
& VLA-JEPA~\citep{vlajepa} & -- & -- & -- & -- & 65.2 & 57.3 & -- \\
& JEPA-VLA~\citep{jepavla} & 73.5 & 17.7 & -- & -- & -- & -- & -- \\
& WoG~\citep{wog} & -- & -- & -- & -- & 69.4 & 63.5 & -- \\
\bottomrule
\end{tabular}

\begin{tablenotes}[flushleft]
\footnotesize
\item Methods are grouped by the manner in which world modeling is incorporated into policy learning, rather than by publication year alone.
\item \textbf{RT-A} and \textbf{RT-B} denote the two RoboTwin evaluation settings, corresponding to the simple setting with non-randomized testing environments and the harder setting with randomized environments, respectively.
\item \textbf{C-A} and \textbf{C-D} denote the CALVIN ABCD and ABCDD protocols, respectively.
\item \textbf{S-G}, \textbf{S-W}, and \textbf{S-O} denote SIMPLER-style results on Google Robot, WidowX, and other reported setups, respectively.
\item ``--'' indicates that no directly reported number was found under the corresponding protocol. Entries of the form ``a/b'' denote two reported protocol variants in the original source.
\end{tablenotes}
\end{threeparttable}
\end{table}

Building on the previous discussions of evaluation protocols and training data, we briefly summarize representative results on common downstream manipulation benchmarks. Because evaluation criteria for embodied world models are often benchmark-specific, we focus on task success rate and closely related completion metrics, which are the most widely reported and directly comparable indicators of downstream performance.

Tables~\ref{tab:all_libero_total} and~\ref{tab:all_other_total} collect representative results from recent embodied world-model and world-model-related methods. For clarity, we group methods by how world modeling is integrated with policy learning, including decoupled pipelines, shared-backbone designs, mixture-based architectures, unified VLA-style formulations, and latent-space world-model variants. Although these categories are not strictly exclusive, they provide a useful high-level view of design differences.

Table~\ref{tab:all_libero_total} focuses on the LIBERO~\citep{liu2023libero} standard 4-suite setting. We retain the breakdown into Spatial, Object, Goal, and Long suites, since methods with similar averages can still differ substantially across subsets. Table~\ref{tab:all_other_total} complements this with results on RoboTwin 2.0~\citep{chen2025robotwin}, CALVIN~\citep{mees2022calvin}, and SIMPLER-style~\citep{li2025evaluating} benchmarks. These evaluations are more heterogeneous in embodiment and protocol, so they are less suitable for strict ranking but still useful for revealing cross-benchmark variation.

Several patterns are worth noting. First, strong results are not limited to a single architectural paradigm: competitive performance appears across decoupled, shared-backbone, unified, mixture-based, and latent predictive designs. This suggests that the utility of world modeling for embodied control is not tied to one specific implementation. Second, the LIBERO breakdown shows that long-horizon manipulation remains a key differentiator. While many methods already perform strongly on Spatial and Object suites, larger drops are more common on Goal and especially Long suites, where success depends more on sustained, action-grounded consistency over extended trajectories.

Results on RoboTwin, CALVIN, and SIMPLER-style benchmarks further support this point, while also highlighting stronger benchmark dependence. Compared with LIBERO, these settings are more fragmented, and strong performance on one benchmark does not necessarily transfer to another. This suggests that current embodied world models are still sensitive to differences in embodiment, action space, task composition, and evaluation protocol.

Overall, these results suggest three conclusions. First, embodied world models already show strong practical utility on standard downstream manipulation benchmarks. Second, high performance can emerge from multiple design paradigms, indicating that photorealistic video generation is not necessary for effective embodied control. Third, the main remaining challenges lie in long-horizon robustness, cross-benchmark generalization, and the lack of standardized reporting across platforms.

% 
% Optional 

\section{Challenges and Future Directions}
\label{sec:challenges}
\todo[inline, color=mybrown]{Assigned to: Jindou Jia , Jianfei Yang}

%\jjd{drafted by jjd 02/26}
%\jjd{revised by jjd 02/28}
%\jjd{revised by jjd 03/10}

Despite the promise of world models for robot learning, reliable deployment in complex embodied tasks remains limited by several core challenges beyond simple scaling. Current systems must address causal conditioning gaps in action-dependent dynamics, efficiency bottlenecks in training and inference, limited integration of non-visual sensory feedback, and the lack of standardized evaluation focused on functional utility rather than visual realism. Another important frontier is symbolic and structured abstraction: while pixel- or latent-space prediction is powerful, long-horizon reasoning may require object-centric, relational, or rule-like structure that provides a more compact interface for planning and control. In this section, we discuss these challenges and outline future directions toward reliable, efficient, and actionable world models for embodied agents.

\subsection{Causal Conditioning Gaps}

Current VLA frameworks often couple world models with inverse dynamics~\citep{lingbotva,dreamzero,gigabrain}, using future-state prediction to regularize policy learning. However, a causal misalignment can arise when the predicted future is conditioned more strongly on historical context or task intent than on the specific pending robot action. In such cases, the world model may generate futures that are semantically plausible or intention-consistent, but not necessarily faithful to the physical consequences of the candidate action. This limits its usefulness for precise closed-loop control, where the key requirement is not only to predict a likely future, but to predict how the future changes under the robot's own intervention.

The technical bottleneck is weak action conditioning: many predictive world-model objectives are trained mainly from observation history and task intent, so their futures can be plausible without being causally tied to the robot action to be executed. To reduce this mismatch, WorldVLA~\citep{rynnvla002} adopts implicit unified training strategies that couple future-state prediction with action generation, encouraging more policy-aligned predictive dynamics.

\subsection{Efficiency Bottlenecks} 
%\gen{drafted on 03/11, revised on 03/30}
% with studies reporting up to 3$\times$ longer training times~\citep{wu2026vlanext}

World-model-based policies are far more computationally intensive than VLA models, especially during both training and inference. This overhead arises because models either jointly predict future videos and actions or require fine-tuning before policy learning, making adaptation costly due to large model sizes and complex environment dynamics. Parameter-efficient strategies, like lightweight adapters, can mitigate this by keeping the base model largely frozen.
Efficiency issues also appear at inference, particularly for diffusion-based video prediction, where iterative denoising causes high latency. Recent approaches like Mimic Video~\citep{mimicvideo} and LingBot-VA~\citep{lingbotva} mitigate this via partial denoising. These methods prioritize motion dynamics over fine-grained visual details, capturing essential cues for decision-making without the cost of full reconstruction.

More fundamentally, recent approaches rethink world models entirely. Latent-space models, such as LeWorldModel~\citep{leworldmodel}, reduce training and inference costs by focusing on predictive representations rather than full high-dimensional generation. Emerging paradigms, like Fast-WAM~\citep{fastwam}, further decouple world modeling from deployment, using it only for training to enhance representations while eliminating it at inference.

\subsection{Multi-Modal Perception Bottlenecks}

Current world models excel in visual synthesis but remain decoupled from the physical dynamics of real-world interaction. Relying predominantly on vision and proprioception fails to capture unobservable properties such as friction, stiffness, and contact stability. To resolve these, integrating haptic sensing and force feedback~\citep{mode_vla, tactile_vla} is indispensable for providing ground-truth interaction signals. Recent visuo-tactile models~\citep{higuera2026vtwm, zheng2026omnivta} have begun addressing this by learning joint latent representations to enhance robustness in contact-rich tasks.

A significant architectural challenge lies in aligning asynchronous signals with divergent frequencies and dimensions. While tactile sensors capture high-frequency transient events, their low-dimensional signals are often diluted or overwhelmed by high-dimensional visual features during joint latent optimization~\citep{chen2025multi}. Effectively balancing these heterogeneous inputs is essential for preventing visual dominance and ensuring that sparse visual semantics are fused with dense physical feedback, a critical step toward physics-aware robotic intelligence.

%\subsection{Classical Control Integration} 

%World Models can serve as the environment's forward dynamics in the context of control theory, enabling the agent to predict the subsequent state given the current state and a candidate action \citep{hansen22a, hansen2024tdmpc2}. Recent advancements increasingly integrate these models with Model Predictive Control (MPC) to facilitate proactive planning. By utilizing the world model as the transition backbone, the agent can optimize an action sequence $\{a_{t:t+H}\}$ over a finite horizon $H$ to minimize a cumulative cost $\sum_{k=0}^{H} \mathcal{C}(s_{t+k}, a_{t+k})$. This synergy allows the robot to perform "imagined" rollouts to evaluate potential future outcomes, effectively bridging the gap between reactive policy execution and long-horizon strategic reasoning. 

%Distinct from classical control theory, which predominantly relies on analytical physics models centered on the robot’s intrinsic kinematics and dynamics, world models offer a more holistic representation. They encapsulate the joint evolution of the agent’s internal state and the external environment's stochasticity. A promising frontier lies in the integration of learned world models with established control principles, such as Lyapunov-based stability or robust control. Fusing the expressive power of neural world models with the formal guarantees of control theory presents a robust pathway toward enhancing the self-adaptive capabilities of robotic systems in non-stationary, open-world environments.

\subsection{Classical Control Integration}

World models serve as forward dynamics for proactive planning via MPC \citep{hansen22a, hansen2024tdmpc2, leworldmodel}. By optimizing action sequences to minimize cumulative costs, agents use imagined rollouts to bridge reactive execution with strategic reasoning. However, a major bottleneck is the massive computational overhead. MPC requires iterative world model rollouts for action optimization, which significantly limits the real-time deployment of high-capacity models in dynamic environments.

Unlike analytical kinematics, world models capture the joint stochastic evolution of both the agent and its environment. A critical frontier lies in reconciling this neural expressivity with formal control guarantees, such as Lyapunov stability or robust control~\citep{jia2024feedback}. Fusing learned dynamics with existing mature control principles, not just MPC, presents a potential pathway toward self-adaptive robotic systems capable of operating in non-stationary, open-world settings.

\subsection{Symbolic Structure Integration}

While this survey has primarily focused on visual and latent world models, symbolic world models provide an important complementary direction. Instead of predicting pixels, they operate over structured states, such as objects, relations, predicates, or occupancy maps, enabling more stable and compositional predictions. A key limitation of pixel-based rollouts is long-horizon error accumulation, which can degrade planning reliability. Symbolic representations mitigate this by abstracting away low-level details and modeling discrete or rule-based transitions, allowing more reliable reasoning over extended horizons. However, they often require suitable abstractions and perception grounding, and can struggle when high-dimensional observations cannot be cleanly mapped into predefined symbols. A promising direction is therefore to build hybrid world models that combine learned perceptual representations with symbolic structure~\citep{exopredicator,visualpredicator,lamp_symbolicwm}. This is compelling because much of the real world is inherently structured: object-centric or relational abstractions learned from data, together with symbolic constraints in generative models, may offer a principled path toward scalable and reliable long-horizon world modeling.

% specific for world action model. lightweight.
\subsection{Open Challenges in Evaluation Metrics}
%\ant{draft by 4/2}

Another challenge for embodied world models is the absence of a widely accepted evaluation metric. Unlike conventional video generation, where perceptual fidelity is often central, embodied world models are ultimately judged by their functional value for decision making~\citep{shang2026worldarena,zhang2025world}. A model may produce visually plausible futures yet still fail to preserve action-conditioned dynamics, causal consistency, or controllability, all of which are critical for policy learning and closed-loop execution. Conversely, limited visual realism does not necessarily preclude utility for planning or policy evaluation~\citep{quevedo2025worldgym}. As a result, evaluation remains inherently multi-dimensional, spanning predictive quality, downstream control utility, and physical executability~\citep{fan2026wow}, and current comparisons are still fragmented across benchmarks and protocols.

A key direction is therefore to develop function-aware evaluation frameworks that better reflect the intended role of the world model. Rather than relying on appearance-driven scores alone, future metrics should jointly assess predictive realism, action sensitivity, long-horizon consistency, and control utility. A practical goal is to establish a compact set of standardized metrics, such as task success, policy-ranking fidelity, and executability-oriented diagnostics, to enable more consistent comparison across tasks and embodiments and to distinguish visually plausible models from truly actionable ones.

% \subsection{Few-Demonstration for Robot Learning}

\clearpage
\newpage
\bibliographystyle{assets/plainnat}
\bibliography{paper}

\end{document}